\documentclass[runningheads]{llncs}
\renewcommand{\lastandname}{\unskip,}

\usepackage{accv}

\usepackage{accvabbrv}

\usepackage{graphicx}
\usepackage{booktabs}

\usepackage[accsupp]{axessibility}  % Improves PDF readability for those with disabilities.

\usepackage[pagebackref,breaklinks,colorlinks,citecolor=accvblue]{hyperref}
\usepackage{graphicx}
\usepackage{booktabs}
\usepackage{adjustbox}
\usepackage{comment}
\usepackage{multirow}
\usepackage{booktabs}
\usepackage{subcaption}
\usepackage[table]{xcolor}

\usepackage{orcidlink}

\begin{document}

\title{NBAvatar: Neural Billboards Avatars \\ with Realistic Hand-Face Interaction}

\titlerunning{NBAvatar}

\author{David Svitov\inst{1,2}\orcidlink{0009-0009-9116-0416} \and
Mahtab Dahaghin\inst{1,2}\orcidlink{0009-0001-5559-4884} \and
Pietro Morerio\inst{2}\orcidlink{0000-0001-5259-1496} \and
Alessio Del Bue \inst{2}\orcidlink{0000-0002-2262-4872}}

\authorrunning{D.~Svitov et al.}

\institute{Università degli Studi di Genova, Genoa, Italy \and
Istituto Italiano di Tecnologia (IIT), Genoa, Italy\\
\email{\{david.svitov, mahtab.dahaghin, pietro.morerio, alessio.delbue\}@iit.it}}

\maketitle

\begin{abstract}
  We present NBAvatar - a method for realistic rendering of head avatars handling non-rigid deformations caused by hand-face interaction. To this end, we introduce a novel hybrid implicit-explicit representation for animated avatars by combining the training of explicit oriented planar primitives with implicit neural rendering. Such a combination of representations in the end-to-end pipeline enables NBAvatar to handle temporally and pose-consistent geometry, along with fine-grained appearance details provided by the neural rendering technique. To enable joint optimization of different representations we propose a geometry-aware training scheme that allows our hybrid representation to surpass existing approaches in terms of novel-view and novel-pose rendering quality. 
  Specifically, NBAvatar achieves up to 53\% LPIPS reduction compared to Gaussian-based avatar methods, while also improving PSNR and SSIM, and achieves higher structural similarity compared to the state-of-the-art hand-face interaction method InteractAvatar. \\
  Project page: \href{https://david-svitov.github.io/NBAvatar_project_page}{https://david-svitov.github.io/NBAvatar\_project\_page}

\end{abstract}

\section{Introduction}
\label{sec:intro}

\begin{figure}[tb]
  \centering
  \includegraphics[width=12.0cm]{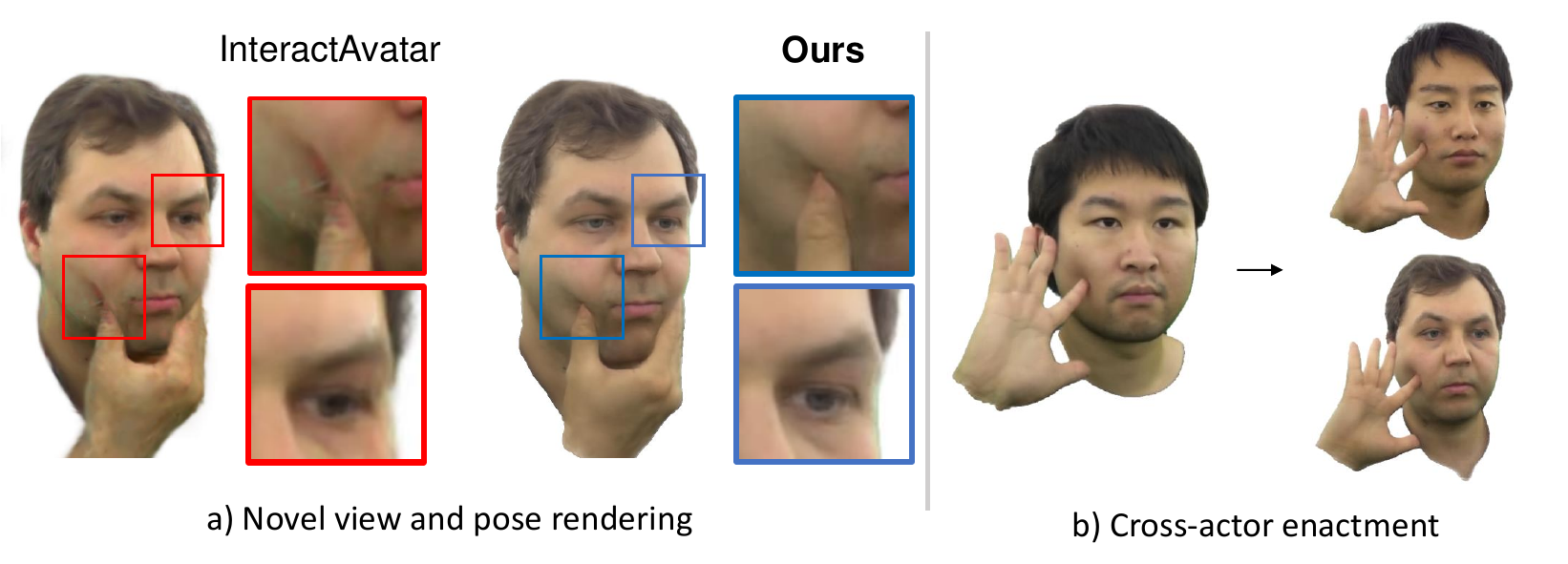}
  %{images/teaser/teaser_enacment_1.png}
  \caption{\textbf{Novel view synthesis.} We present the method for photo-realistic novel-view rendering of non-rigid hand-face interaction scenarios for human avatars. a) Our method achieves photo-realistic rendering quality for novel views and poses, surpassing state-of-the-art InteractAvatar~\cite{chen2025interactavatar} results. b) The method enables cross-actor enactment by transferring hand and face poses across different subjects.}
  \label{fig:teaser}
\end{figure}

The interaction between hands and face is an important source of information in human communication \cite{dimond1984face, mueller2019self} and therefore is of paramount importance for telepresence and virtual reality applications. Such applications rely on rendering of the dynamic hand and face, but struggle to realistically represent their interaction. Modern methods concentrate solely on head \cite{gecer2019ganfit, feng2021learning, lombardi2018deep, gafni2021dynamic, zheng2022avatar, zielonka2023instant, qian2024gaussianavatars, shao2024splattingavatar} or hand \cite{chen2023hand, kalshetti2024intrinsic, guo2023handnerf, mundra2023livehand, guo2025handnerf++, corona2022lisa, pokhariya2024manus} rendering and leave their mutual interaction out of scope. While a few methods \cite{chen2025interactavatar, he2025capturing} attempt to handle the rendering of hand-face interaction, they often struggle with artifacts that limit their applicability. In this work, we address the task of photo-realistically rendering of face-hand interactions (\cref{fig:teaser} (a)), accounting for non-rigid face deformation and color changes under both novel viewpoints and novel poses - the latter being essential for avatar animation and self and cross-reenactment (\cref{fig:teaser} (b)) in telepresence.

Head avatar methods \cite{thies2019deferred, gecer2019ganfit, zielonka2023instant, qian2024gaussianavatars, shao2024splattingavatar} rely on different rendering approaches to achieve photo-realistic novel-view synthesis. 
NeRF-based methods \cite{gafni2021dynamic, grassal2022neural, kirschstein2023nersemble, zielonka2023instant, mildenhall2020nerf} provide view-consistent rendering but suffer from slow inference, even with acceleration techniques \cite{muller2022instant, zielonka2023instant}. 
DNR~\cite{thies2019deferred} achieves real-time speed by deferring neural processing to screen space, but tends to produce view-dependent overfitting artifacts due to limited underlying mesh detail.
The breakthrough in human avatar representation happened with the appearance of 3D Gaussian Splatting (3DGS)~\cite{kerbl20233d}. Modern head avatar methods \cite{qian2024gaussianavatars, shao2024splattingavatar, saito2024relightable, li2024uravatar} utilize 3DGS to handle realistic animated rendering \cite{qian2024gaussianavatars, shao2024splattingavatar} and lighting effects \cite{saito2024relightable, li2024uravatar}. While the head avatar's rendering quality has significantly improved, non-rigid face deformations and hand-face interactions remain underrepresented. %in the existing literature.
Pioneering approaches \cite{shimada2023decaf, wu2024dice, he2025capturing} in hand-face interaction aimed to predict face mesh deformation caused by hand touch, but did not target rendering realism. 
Decaf \cite{shimada2023decaf} first solved the task of predicting face mesh non-rigid deformations by introducing an encoder-decoder model to predict hand-face interaction area and mesh offsets from monocular video. 
The pipeline proposed in \cite{he2025capturing} is among the first to render the resulting mesh using a rendering network, but it produces excessively blurred results. The recent work InteractAvatar \cite{chen2025interactavatar} represents appearance using 3DGS and 
interaction-aware MLPs predicting pose-driven color and position offsets.
While InteractAvatar successfully simulates self-cast shadows and face deformations, it inherits 3DGS-specific artifacts, blurry facial textures and Gaussians protruding at the body boundary, that limit its realism (\Cref{fig:teaser}(a)).

%While existing approaches rely solely on implicit or explicit representation, we introduce an efficient combination of these paradigms. 
In this work, we propose a novel \textbf{Neural Billboard} primitive that represents the scene as a synergy of implicit and explicit learnable representations optimized within a hybrid end-to-end pipeline. Neural Billboards use implicit neural fields encoded with neural textures \cite{thies2019deferred} along the surface of explicit 3D-oriented billboards \cite{svitov2025billboard}. While textured billboard primitives \cite{svitov2025billboard} offer surface-aligned explicit geometry, their RGB parameterization limits adaptivity in dynamic contact regions. Conversely, neural textures \cite{thies2019deferred} enable view- and pose-dependent effects but were originally designed for fixed mesh parameterizations and are unstable when naively applied to independently optimized geometric primitives. Neural Billboards successfully adapts the neural texture paradigm to learnable planar primitives describing the avatar surface.

Combining implicit neural textures with explicit 3D planar primitives in an end-to-end pipeline is a non-trivial task: without additional constraints, the optimization becomes ill-posed, as geometry and appearance compete to explain changes in silhouette and shading. To ensure convergence of our hybrid avatar representation, we propose a training technique that disentangles geometry and appearance during optimization by introducing an additional intermediate segmentation optimization. To summarize, our contributions are as follows:
\begin{itemize}
    \item We introduce Neural Billboards, a hybrid implicit-explicit representation reconciling surface-aligned planar primitives with deferred neural rendering.
    \item We propose a geometry-aware training scheme with intermediate silhouette supervision that stabilizes joint optimization of spatial and neural texture parameters for hybrid representations.
    \item We demonstrate that explicit geometry representation can be optimized along with an implicit appearance model in an end-to-end joint and disentangled way, rather than handled in separate stages. 
\end{itemize}

\section{Related Works}
\label{sec:related_works}

\textbf{Rendering Methods.}
%Deferred Neural Rendering (DNR)~\cite{thies2019deferred} for the first time achieved photorealistic avatar rendering by decoupling geometry from appearance: a mesh is rasterized with per-vertex learned neural feature vectors, and the resulting screen-space feature image is decoded into a photorealistic output by a convolutional network. This design enables high visual fidelity at real-time inference speed, since the computationally expensive neural processing is deferred to a single image-space pass rather than applied per-point in 3D\cite{mildenhall2020nerf}. However, the rendering quality remains tied to the resolution and accuracy of the underlying mesh, as the feature maps are defined on mesh vertices.

Deferred Neural Rendering (DNR)~\cite{thies2019deferred} introduced photorealistic avatar rendering by decoupling geometry from appearance. A mesh with learned per-vertex features is rasterized into a screen-space feature map and decoded into a photorealistic image by a convolutional network. This enables high visual fidelity at real-time speed by performing neural processing once in image space rather than per-point in 3D~\cite{mildenhall2020nerf}. However, rendering quality remains limited by the resolution and accuracy of the underlying mesh.

Neural Radiance Fields (NeRF)~\cite{mildenhall2020nerf} represent scenes as continuous volumetric functions evaluated along camera rays, enabling high-quality novel view synthesis but incurring substantial computational cost due to dense MLP evaluation. Subsequent acceleration techniques~\cite{muller2022instant, fridovich2022plenoxels, chen2022tensorf} improved efficiency but retained volumetric ray sampling, limiting real-time performance.

%Neural Radiance Fields (NeRF)~\cite{mildenhall2020nerf} represented the scene as continuous volumetric functions queried via sampling along the camera ray instead of relying on the mesh. However, the dense per-ray MLP evaluations made rendering significantly slower than DNR rasterization. Instant-NGP~\cite{muller2022instant} accelerated both training and inference using a multi-resolution hash encoding, but the volumetric formulation still incurred overhead caused by dense MLP evaluation. Subsequent works sought to eliminate the costly MLP evaluation altogether by representing radiance fields explicitly in structured grids or tensor factorizations. Plenoxels~\cite{fridovich2022plenoxels} replaced neural networks with sparse voxel grids storing spherical harmonics coefficients, while TensoRF~\cite{chen2022tensorf} factorized volumetric features into low-rank tensor components. Although these approaches significantly improved efficiency, they retained volumetric ray sampling and thus remained computationally demanding for real-time avatar rendering.

3D Gaussian Splatting (3DGS)~\cite{kerbl20233d} reconciled quality and speed by returning to explicit primitives: the scene is represented as a collection of %anisotropic 
3D Gaussians that are projected 
%and $\alpha$-blended via tile-based rasterization
into the image-space, enabling real-time rendering with quality competitive to NeRF. 
%However, a large number of overlapping primitives are needed to capture fine texture detail. 
2D Gaussian Splatting (2DGS)~\cite{huang20242d} proposed planar Gaussians that better align with object surfaces, improving geometric accuracy and enabling more reliable mesh extraction. 
To reduce the number of primitives required for fine texture detail, Billboard Splatting (BBSplat)~\cite{svitov2025billboard} replaces each 2D Gaussian with a textured planar primitive that carries learnable RGB textures and alpha maps.

%Our method draws inspiration from both deferred neural rendering and billboard-based splatting. \textcolor{red}{**TO FILL**}

\textbf{Head Avatars.} 
To model realistic, drivable head avatars, early mesh-based approaches~\cite{blanz2023morphable, blanz2003reanimating, paysan20093d} used parametric 3D Morphable Models (3DMM), such as FLAME~\cite{li2017learning}, fitted to facial images, but were limited to low-frequency appearance due to coarse per-vertex attributes. Texture-based extensions~\cite{gecer2019ganfit, feng2021learning, lombardi2018deep} improved visual fidelity by attaching RGB textures to the mesh surface, yet remained constrained by the fixed resolution and topology of the underlying geometry. 
%A key breakthrough came with methods that decoupled geometry from appearance: 
DNR~\cite{thies2019deferred} proposed to utilize learned neural feature maps instead of hand-crafted textures to represent out-of-mesh details via deferred rendering. 
%Based on this technique, VariTex~\cite{buhler2021varitex} learned a variational latent space of neural face textures enabling novel identity sampling, and 
Subsequent works~\cite{buhler2021varitex, grassal2022neural, bashirov2024morf} further explored neural texture representations for face avatars.
%Further works \cite{buhler2021varitex, grassal2022neural, bashirov2024morf} extended this idea; NHA~\cite{grassal2022neural} proposed combining a 3DMM with feed-forward networks to predict both vertex offsets and view-dependent textures. 
While such models enable realistic rendering, they often produce overfitting artifacts on the out-of-mesh details.

The NeRF~\cite{mildenhall2020nerf} representation enabled view-consistent realistic rendering and shifted the field toward fully implicit representations. NerFace~\cite{gafni2021dynamic} and IMavatar~\cite{zheng2022avatar} conditioned NeRFs on 3DMM parameters to enable controllable synthesis from monocular video, with IMavatar further improving generalization to unseen expressions through learned implicit blendshapes and skinning fields. 
%However, these methods suffered from prohibitively slow volumetric rendering. Subsequent works attempted to alleviate this limitation either by introducing efficient spatial encodings and acceleration structures, such as INSTA~\cite{zielonka2023instant} and AvatarMAV~\cite{xu2023avatarmav}, or by reformulating the radiance representation itself, as in tri-plane-based 3D-aware models like EG3D~\cite{chan2022efficient}. Nevertheless, the reliance on volumetric ray sampling or dense feature querying remains computationally demanding for real-time avatar applications.
However, these methods suffered from prohibitively slow volumetric rendering. INSTA~\cite{zielonka2023instant} and AvatarMAV~\cite{xu2023avatarmav} partially addressed this by embedding radiance fields within efficient spatial data structures, but the fundamental overhead of volumetric ray sampling remained.

3DGS~\cite{kerbl20233d} resolved the speed-quality trade-off by introducing explicit Gaussian primitives to represent the scene. PointAvatar~\cite{zheng2023pointavatar} pioneered deformable point-based head avatars from monocular video, while GaussianAvatars~\cite{qian2024gaussianavatars} and SplattingAvatar~\cite{shao2024splattingavatar} established the hybrid mesh-Gaussian paradigm by binding Gaussian kernels to FLAME~\cite{li2017learning} template mesh surfaces, enabling controllable animation with real-time rendering.
%GaussianAvatars fixes each Gaussian to a specific triangle, while SplattingAvatar allows Gaussians to migrate across faces during optimization. 
More recently, CloseUpAvatar~\cite{svitov2025closeupavatar} replaced Gaussians with textured billboard primitives~\cite{svitov2025billboard}, achieving high-fidelity rendering across a wider range of camera distances, %than Gaussian-based counterparts
but assuming a static appearance of avatars. Despite these advances, all of the above methods focus exclusively on head modeling and do not address the dynamics of hand-face interaction.

%In our work, we adopt 3DMM templates, FLAME~\cite{li2017learning} for the face and MANO~\cite{romero2022embodied} for the hand, \textcolor{red}{**TO FILL**}

\textbf{Hand-Face Interaction.}
While significant progress has been made in reconstructing photorealistic heads~\cite{gecer2019ganfit, feng2021learning, lombardi2018deep, gafni2021dynamic, zheng2022avatar, zielonka2023instant, qian2024gaussianavatars, shao2024splattingavatar} and hands~\cite{chen2023hand, kalshetti2024intrinsic, guo2023handnerf, mundra2023livehand, guo2025handnerf++, corona2022lisa, pokhariya2024manus} independently, the photo-realistic rendering of the non-rigid deformations and color change caused by hand-face contact remains underrepresented in the literature.
%the non-rigid deformations that arise when these two body parts come into contact, such as skin compression, shadowing, and tissue displacement, remain largely unaddressed, despite being essential for conveying natural nonverbal communication in avatars.
%The foundational work in this space 
One of the first works in the field
is Decaf~\cite{shimada2023decaf}, which introduced a 
%first 
monocular RGB method for jointly tracking hand-face interactions and estimating facial deformations using a position-based dynamics (PBD)~\cite{muller2007position} simulator. %with anatomically motivated stiffness values. 
DICE~\cite{wu2024dice} extended this to end-to-end estimation via a Transformer architecture. NePHIM~\cite{wagner2025nephim} proposed a volumetric simulation incorporating temporal effects, anatomical constraints, and skin-pulling dynamics, approximated by lightweight neural networks for real-time performance. However, all of these methods operate solely at the mesh level, without addressing photorealistic rendering.
%While producing more natural deformations than PBD-based methods, NePHIM operates on mesh templates and does not model appearance.
He~et~al\onedot~\cite{he2025capturing} proposed to learn a PCA basis for hand-induced facial deformations, achieving physically plausible reconstructions, but the focus remains on geometry rather than high-fidelity appearance.
%addressed the monocular setting by learning a PCA basis for hand-induced facial deformations from iPhone video, achieving physically plausible reconstructions without multi-view capture, but the PCA basis constrains the deformation space and the focus remains on geometry rather than high-fidelity appearance.

InteractAvatar~\cite{chen2025interactavatar}, directly extended the hybrid mesh-Gaussian paradigm \cite{qian2024gaussianavatars, shao2024splattingavatar} %of GaussianAvatars~\cite{qian2024gaussianavatars} and SplattingAvatar~\cite{shao2024splattingavatar} 
to handle the realistic rendering of hand-face interactions. While 
%both baselines 
3DGS-based approaches produce compelling face reconstructions, but they lack mechanisms for pose-dependent appearance dynamics or non-rigid inter-body deformation, resulting in 
%over-smoothed hands and rigid, 
artifact-prone contact regions. InteractAvatar addresses this by augmenting the framework with MLPs to control the appearance of the dynamic hand and interaction region. These MLPs predict offsets on the MANO~\cite{romero2022embodied} hand mesh and PBD-initialized facial deformation area based on the hand pose, face expression, and global geometric features calculated by a PointNet++~\cite{qi2017pointnet++}.
%a Dynamic Gaussian Hand that uses pose-conditioned MLPs to predict per-Gaussian geometry and appearance offsets on the MANO~\cite{romero2022embodied} mesh, and a hand-face interaction module that refines PBD-initialized facial deformations via an MLP conditioned on hand pose, face expression, and global geometric features from a PointNet++. 

While InteractAvatar achieves high rendering quality, it remains subject to 3DGS-related artifacts caused by the Gaussians' displacements. These artifacts are particularly noticeable along the outline of the avatar, resulting in a frosty pattern, as well as general blurriness of the face. Our approach circumvents these issues by combining explicit and implicit representations, such as Billboards~\cite{svitov2025billboard} and DNR~\cite{thies2019deferred} neural textures to handle high-frequency details and pose-dependent effects in rendering.

\section{Method}
\label{sec:method}

\subsection{Preliminaries: Billboard Splatting}
Billboard Splatting was introduced in~\cite{svitov2025billboard} as a texture-based alternative to 3D Gaussian Splatting (3DGS)~\cite{kerbl20233d} and its 2D variant (2DGS)~\cite{huang20242d}. The method represents a scene using textured surfels (billboards) instead of volumetric Gaussians. Unlike other texture-driven primitives~\cite{weiss2024gaussian, rong2025gstex, song2024hdgs, xu2024supergaussians}, the proposed representation jointly optimizes both RGB and opacity textures while maintaining real-time rendering performance.

The spatial parametrization follows the 2DGS formulation for position and orientation, while augmenting each primitive with learnable texture maps. Specifically, each billboard is defined as $\{\mu_i, s_i, r_i, \textrm{SH}_i, T_i^\textrm{RGB}, T_i^\alpha\}$, where $\mu_i \in \mathbb{R}^3$ denotes the 3D center, $s_i \in \mathbb{R}^2$ the anisotropic 2D scale, and $r_i \in \mathbb{R}^4$ the quaternion encoding its rotation. The term $\textrm{SH}_i \in \mathbb{R}^{4 \times 3}$ represents spherical harmonics coefficients, while the values $\{T_i^\textrm{RGB}, T_i^\alpha\}$ correspond to learnable textures of size $S^T \times S^T$ texels for color and opacity.

For rendering, BBSplat employs an explicit ray–splat intersection algorithm~\cite{sigg2006gpu} to compute billboard-space coordinates $(u,v)$ corresponding to a screen-space pixel $(x,y)$. The normalized coordinates $(u,v)\in[-1,1]$ are then mapped to the discrete texture grid and used to sample values from $T_i^\textrm{RGB}$ and $T_i^\alpha$.

%\subsection{Preliminaries: Billboard Splatting}
%Billboard splatting was proposed in \cite{svitov2025billboard} as an alternative to 3DGS~\cite{kerbl20233d} and 2DGS~\cite{huang20242d}, and utilizes textured surfels (billboards) for 3D scene representation. In contrast with other texture-based primitives~\cite{weiss2024gaussian, rong2025gstex, song2024hdgs, xu2024supergaussians}, billboards allow both RGB and opacity texture training and enable real-time inference speed.  Billboards follow the 2DGS parametrization for location and orientation in space and additionally store learnable RGB and alpha textures. The parametrization of the billboards is as follows: $\{\mu_i, s_i, r_i, \textrm{SH}_i, T_i^\textrm{RGB}, T_i^\alpha\}$, which corresponds to the  position $\mu_i \in \mathbb{R}^3$, 2D scale $s_i\in \mathbb{R}^2$, quaternion rotation $r_i \in \mathbb{R}^4$, and spherical harmonics $\textrm{SH}_i \in \mathbb{R}^{4 \times 3}$ of the primitive. The values $\{T_i^\textrm{RGB}, T_i^\alpha\}$ correspond to learnable textures of size $S^T \times S^T$ texels for color and opacity.

%During rasterization, BBSplat uses the explicit ray-splat intersection algorithm~\cite{sigg2006gpu} to find the corresponding billboard coordinate $(u, v)$ for a given screen coordinate $(x, y)$. The resulting $(u, v) \in [-1, 1]$ coordinates then rescale for the predefined texture size $S^T$ and sample values from $T_i^\textrm{RGB}$ and $T_i^\alpha$.

\subsection{Neural Texture Primitives}

\begin{figure}[tb]
  \centering
  \includegraphics[width=12.0cm]{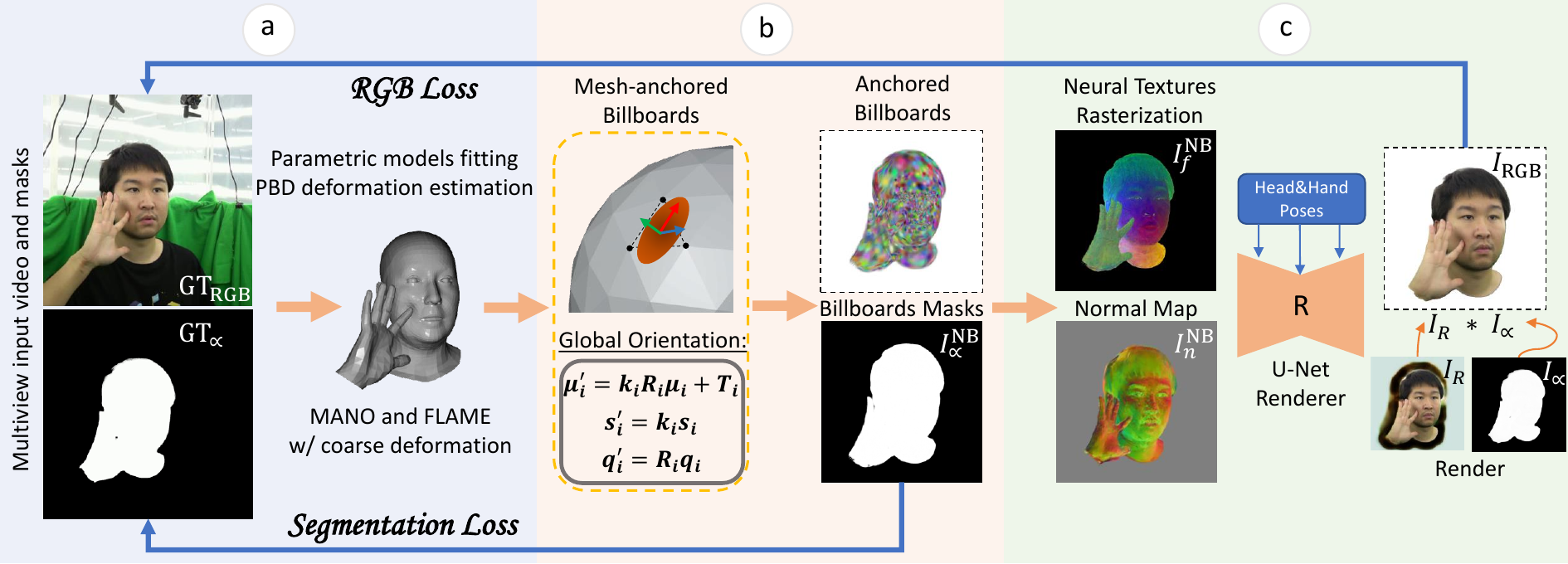}
  \caption{\textbf{Method description.} a) For multi-view input video, we fit FLAME~\cite{li2017learning} and MANO~\cite{romero2022embodied} parametric models and reconstruct coarse face deformation with position-based dynamics (PBD)~\cite{muller2007position}. b) Then, for each polygon in the mesh, we anchor Neural Billboards by computing its orientation as offsets relative to the polygon. We rasterize Neural Billboards into a 6-channel image and a corresponding alpha map. c) Finally, we transform this rasterization to RGB with the U-Net renderer $R$ while regularizing the alpha map to fit the ground truth silhouette.}
  \label{fig:main_scheme}
\end{figure}

\begin{figure}[tb]
  \centering
  \includegraphics[width=12.0cm]{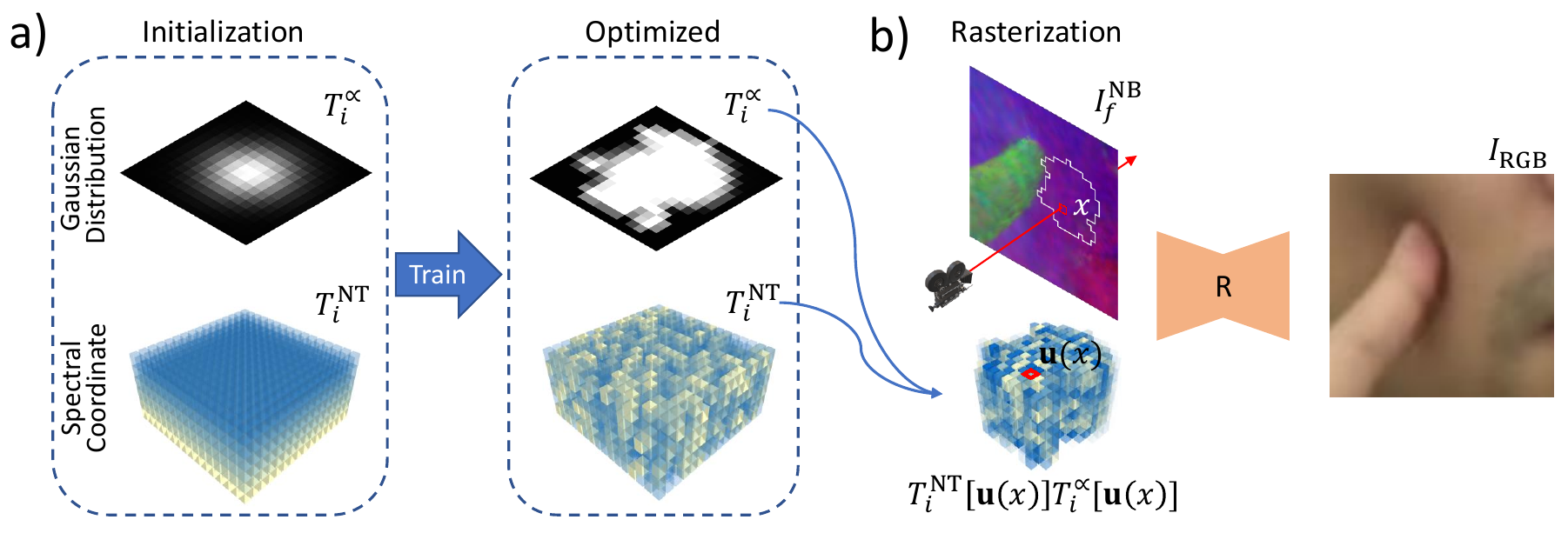}
  \caption{\textbf{Neural billboards rasterization.} a) We initialize the alpha texture $T_i^\alpha$ with a Gaussian distribution, and fill neural texture $T_i^\textrm{NT}$ with corresponding spectral coordinates on the mesh surface. During training, we optimize both textures with gradients from the rasterizer. b) We rasterize screen point $x$ by accumulating corresponding texture values along the ray. The resulting $I_{f}^\textrm{NB}$ transforms to RGB with a trainable renderer $R$.
  }
  \label{fig:features}
\end{figure}

In this work, we propose to combine explicit Billboard-based representation \cite{svitov2025billboard} with implicit Neural Textures (NT) introduced in DNR~\cite{thies2019deferred} as shown in \Cref{fig:main_scheme} in a single end-to-end pipeline. We utilize the explicit representation to describe the avatar's geometry, while implicit NTs handle view- and pose-dependent color effects. The combination of different representations unlocks drivable avatars with photo-realistic detail and real-time rendering. To this end, we introduce a novel \textbf{Neural Billboard} planar primitive that is parametrized with learnable position and orientation $\{\mu_i, s_i, r_i\}$, while using a trainable feature map $T_i^\textrm{NT}$ as a texture and $T_i^\alpha$ alpha map to define the shape. Thus, the final parametrization of the Neural Billboards is $\{\mu_i, s_i, r_i, T_i^\textrm{NT}, T_i^\alpha\}$. We train our Neural Billboard representation for an avatar in an end-to-end manner where $T_i^\textrm{NT}$ and $T_i^\alpha$ are optimized with gradients from the image space losses. While $T_i^\textrm{NT}$ learns feature values, $T_i^\alpha$ is responsible for learning visibility of the planar primitives as shown in \Cref{fig:features}.

During training and inference, each Neural Billboard utilizes neural texture $T_i^\textrm{NT}$ with 6 channels. We start by initializing these textures with spectral coordinates following StylePeople~\cite{grigorev2021stylepeople}. Then we rasterize Neural Billboards into a 6-channel feature map $I_{f}^\textrm{NB}$ and a 1-channel opacity map $I_\alpha^\textrm{NB}$ by accumulating neural texture values along the camera rays:

\begin{equation}
    c(x) = \sum_{i=1} T_i^\textrm{NT}[\textbf{u}(x)] T_i^\alpha[\textbf{u}(x)] \prod_{j=1}^{i-1} (1 - T_j^\alpha[\textbf{u}(x)]),
\end{equation}

where $\textbf{u}$ is $UV$-coordinates on the billboard that correspond to the screen space point $x$. The optimization of $T_i^\alpha$ starts with a Gaussian distribution and learns visibility for each planar point $\textbf{u}$. The rasterization also produces a normal map $I_{n}^\textrm{NB}$ that stores the orientation of the visible billboards in each pixel.

We then employ a UNet-based~\cite{ronneberger2015u} model to transform a concatination of 6-channel rasterization $I_{f}^\textrm{NB}$ and 3-channel normal maps $I_{n}^\textrm{NB}$ into an RGB render $I_\textrm{RGB}$ as shown in \Cref{fig:main_scheme} (c). The UNet renderer $R$ produces the final RGB output $I_\textrm{R}$ along with the transparency map $I_\alpha$, which we combine with background color $C_{bg}$ to obtain the final avatar visualization: 

\begin{equation}
I_\textrm{RGB} = I_\textrm{R} I_\alpha + (1 - I_\alpha) C_{bg}. 
\end{equation}

We inject pose information into our convolutional renderer through Adaptive Instance Normalization \cite{huang2017arbitrary} layers conditioned with FLAME and MANO pose parameters. At normalization layers of the network, the rasterized billboard features are rescaled and shifted by a per-channel affine transformation predicted from the current pose. In contrast with per-Gaussian geometry deformation approaches that condition each primitive individually, the neural renderer $R$ introduces interaction-aware inductive bias via spatial feature aggregation with broadcasted global pose embedding. This image-space appearance processing allows the network to handle pose-dependent changes in the rasterized geometry without any explicit interaction-specific mechanism.

We developed a specialized technique to end-to-end train our Neural Billboard representation along with the renderer $R$ that ensures convergence for both billboard orientation and appearance features. To disentangle the training of spatial orientation from neural features, we introduce an additional segmentation loss for $I_\alpha^\textrm{NB}$ that we discuss in \Cref{seq:disentanglement}. 

\subsection{Driving Avatar Animation}

First, we fit FLAME~\cite{li2017learning} and MANO~\cite{romero2022embodied} parametric models to the multi-view input video (\cref{fig:main_scheme} (a)) based on the detected facial keypoints and hand joints. We also utilize the off-the-shelf SAM3~\cite{carion2025sam} model to accurately segment head and hand regions in input frames. We adjusted the SAM3 prompt to capture only areas that can be represented by available parametric models, namely the head and neck and the hand above the wrist. To capture coarse head mesh deformation caused by interaction with the hand, we use position-based dynamics (PBD)~\cite{muller2007position} physics simulation widely used to model complex skin deformations \cite{bender2015position, chentanez2020cloth, sun2024physhand}. 

As shown in \Cref{fig:main_scheme} (b), we then adapt a common approach to drive 3DGS-based avatars \cite{qian2024gaussianavatars, chen2025interactavatar, svitov2024haha} and anchor Neural Billboards to the mesh surface by calculating their global orientation based on the corresponding  $i$-th polygon transformation $\{T_i, R_i, k_i\}$:

\begin{equation}
    \mu_i' = k_i R_i \mu_i + T_i
\end{equation}

\begin{equation}
    s_i' = k_i s_i
\end{equation}

\begin{equation}
    q_i' = R_i q_i
\end{equation}

Thus, the resulting primitives move along with the parametric mesh surface. Multiplying the billboards' orientation by the polygon rotation $R_i$ and scale $k_i$ ensures that planar primitives are surface-aligned in early training. % aligned with the mesh surface in the initial training steps. 

\subsection{Geometry and Appearance Disentanglement}
\label{seq:disentanglement}

Unlike mesh-based neural textures, our planar primitives undergo independent rigid transformations during animation. This creates an optimization ambiguity: high-frequency appearance effects can be absorbed either by neural features or by the spatial drift of billboards. To address this, we introduce intermediate silhouette supervision at the billboard rasterization stage, explicitly constraining the geometry before neural decoding. This training scheme is essential for stabilizing the joint optimization of spatial and feature parameters.

To achieve this, we utilize a segmentation loss at the Neural Billboards rasterization stage to ensure that the billboards are tightly fitted to the ground-truth silhouette. This way, we disentangle rigid avatar geometry from view- and pose-dependent appearance, such as shadows from self occlusions or small soft tissue deformations from the hand-face interaction.  
We optimize Neural Billboard transparency maps $I_\alpha^\textrm{NB}$ with Dice~\cite{milletari2016v} segmentation loss:

\begin{equation}
    \mathcal{L}_{NB} = \lambda_{NB} Dice(I_\alpha^\textrm{NB}, \textrm{GT}_\alpha)
\end{equation}

The use of billboard segmentation loss is essential to avoid noisy floating primitives 
%avoid rendering artifacts 
as shown in the ablation study in \Cref{fig:ablation}. Please note that we employ a segmentation loss only to ensure billboard alignment in the optimization of the explicit part of the pipeline. The renderer segmentation map $I_\alpha$ is optimized implicitly by using random background colors $C_{bg}$ during RGB training.
For the final rendering, we compute MSE on the top 20\% pixels with the highest reconstruction error, encouraging the model to reconstruct the correct color in challenging regions such as the face. To further push the realism we employ perceptual LPIPS~\cite{zhang2018unreasonable} loss and calculate the final loss as follows:

\begin{equation}
    \mathcal{L}_{RGB} = MSE_{hard}(I_\textrm{RGB}, \textrm{GT}_\textrm{RGB}) + \lambda_{lpips} \mathcal{L}_{lpips}
\end{equation}

\section{Experiments}
\label{sec:experiments}

We compared our method against state-of-the-art head avatar approaches GaussianAvatars~\cite{qian2024gaussianavatars} and SplattingAvatar~\cite{shao2024splattingavatar}, and hand-face interaction rendering approach InteractAvatar~\cite{chen2025interactavatar}. We provide quantitative standard objective metrics and qualitative comparisons with baseline approaches on the Decaf~\cite{shimada2023decaf} hand-face interaction multi-view dataset. In this section, we discuss our choice of baselines and datasets and provide a discussion on the results.

%------------------------------------------------------------------

\subsection{Implementation}
\label{sec:implementation}

To train the avatar appearance, we set the perceptual loss $\lambda_{lpips} = 0.1$ and applied it to the entire image for the first 40,000 iterations and to random $256 \times 256$ crops for the rest of the training to learn fine detail. MSE loss with concentration on top 20\% hard pixels (\ie high MSE error) was used throughout training to preserve the subject's correct colors. We set intermediate segmentation loss coefficients for Billboard segmentation $\lambda_{NB} = 0.1$. 

We trained the neural textures and the renderer $R$ end-to-end for 100,000 iterations with a batch size of 1, using the Adam~\cite{kingma2014adam} optimizer for all parameters. We used a learning rate of 0.0005 for textures and 0.001 for the renderer, and set the billboards' orientation parameters as recommended in \cite{svitov2025billboard}. We utilized a standard UNet architecture and adapted it for deferred rendering by adding two separate heads for rendering $I_\textrm{RGB}$ and $I_\alpha$ at $512 \times 512$ resolution, and AdaIN~\cite{huang2017arbitrary} layers for pose conditioning. For training and inference of all subjects, we used a single NVIDIA GeForce RTX 4090 GPU.

We use $16 \times 16$ textures with six channels for neural texture $T_i^\textrm{NT}$ and a single channel for alpha texture $T_i^\alpha$, and anchored a single billboard to each mesh polygon. We also adapted StopThePop~\cite{radl2024stopthepop} per-ray sorting of surfel primitives to support neural textures. Please refer to the supplementary materials for further implementation details.

%------------------------------------------------------------------

\begin{table}[t]
\centering
\caption{\textbf{Quantitative metrics.} We report metrics for novel views and self-enactment (novel pose) rendering on the Decaf~\cite{shimada2023decaf} dataset.}
\label{tab:metrics}
\setlength{\tabcolsep}{5pt}
\begin{tabular}{l|ccc|ccc}
                & \multicolumn{3}{c|}{Novel view} & \multicolumn{3}{c}{Self-enact} \\ \hline
                & PSNR$\uparrow$ & SSIM$\uparrow$ & LPIPS$\downarrow$ & PSNR$\uparrow$ & SSIM$\uparrow$ & LPIPS$\downarrow$ \\ \hline
%InteractAvatar\textsuperscript{\textdagger}  & 29.85 & 0.9335 & 0.0338 & 28.17 & 0.9356 & 0.0396\\
%SplattingAvatar\textsuperscript{\textdagger} & 25.90 & 0.9135 & 0.0584 & 26.18 & 0.9326 & 0.0546\\[10pt] \hline

SplattingAvatar~\cite{shao2024splattingavatar} & 24.67 & 0.9313 & 0.0942 & 24.83 & 0.9399 & 0.0847 \\
GaussianAvatar~\cite{qian2024gaussianavatars}  & 24.95 & 0.9397 & 0.0756 & 25.14 & 0.9479 & 0.0663 \\
\textbf{Ours} & \cellcolor{orange!25}26.08 & \cellcolor{orange!25}0.9484 & \cellcolor{red!25}0.0439 & \cellcolor{orange!25}25.25 & \cellcolor{orange!25}0.9496 & \cellcolor{red!25}0.0479 \\
\textbf{Ours} (MSE) & \cellcolor{red!25}26.55 & \cellcolor{red!25}0.9517 & \cellcolor{orange!25}0.0518 & \cellcolor{red!25}25.73 & \cellcolor{red!25}0.9523 & \cellcolor{orange!25}0.0543 
\end{tabular}
\end{table}

\subsection{Evaluation Setup}
\label{sec:eval_setup}

\textbf{Datasets.} We evaluate hand-face interaction on the Decaf~\cite{shimada2023decaf} dataset, which is the only publicly available dataset providing multi-view video of subjects performing hand-face interactions with calibrated cameras and fitted FLAME and MANO parameters. Following the experiments of InteractAvatar~\cite{chen2025interactavatar}, we use four subjects and train on seven views while holding out one view for novel view evaluation. We preprocess the data by segmenting the head and hand from the image using the off-the-shelf SAM3~\cite{carion2025sam} model, and cropping the images using the tracked bounding boxes of a person before resizing them to $512 \times 512$ resolution.

\textbf{Baselines.}
We compare NBAvatar against three state-of-the-art methods: GaussianAvatar~\cite{qian2024gaussianavatars}, which attaches 3D Gaussians to 3DMM mesh triangles; SplattingAvatar~\cite{shao2024splattingavatar}, which embeds Gaussians on the mesh surface with learnable migration; and InteractAvatar~\cite{chen2025interactavatar}, which extends the mesh-Gaussian paradigm with interaction-aware MLPs for hand-face deformation and appearance. 
Both GaussianAvatar and SplattingAvatar are the latest hybrid Gaussian head avatar methods that anchor Gaussians to the head template mesh. We naively extend their representation to model hands driven by the MANO hand template mesh.

\textbf{Metrics.}
For quantitative evaluation, following prior work, we report three standard image quality metrics: Peak Signal-to-Noise Ratio (PSNR), Structural Similarity Index Measure (SSIM)~\cite{wang2004image}, and Learned Perceptual Image Patch Similarity (LPIPS)~\cite{zhang2018unreasonable}. While PSNR and SSIM indicate pixel-level reconstruction quality and color matching, LPIPS provides a human perception correlated metric.

\begin{table}[t]
\centering
\caption{\textbf{Comparison with the metrics reported in the InteractAvatar~\cite{chen2025interactavatar} paper.} 
%Images are cropped using subject-specific bounding boxes and resized to $512 \times 512$, following~\cite{chen2025interactavatar}. 
We reproduced the evaluation pipeline for our method as closely as possible, and InteractAvatar results are taken directly from the original paper. %Best per-metric results are in \textbf{bold}. $\uparrow$: higher is better, $\downarrow$: lower is better.
}
\label{tab:ia_protocol}

\setlength{\tabcolsep}{5pt}
\begin{tabular}{l|ccc|ccc}
                & \multicolumn{3}{c|}{Novel view} & \multicolumn{3}{c}{Self-enact} \\ \hline
                & PSNR$\uparrow$ & SSIM$\uparrow$ & LPIPS$\downarrow$ & PSNR$\uparrow$ & SSIM$\uparrow$ & LPIPS$\downarrow$ \\ \hline
InteractAvatar  & \cellcolor{red!25}29.85 & 0.9335 & \cellcolor{red!25}0.0338 & \cellcolor{red!25}28.17 & 0.9356 & \cellcolor{red!25}0.0396\\
\textbf{Ours} & 26.08 & \cellcolor{orange!25}0.9484 & \cellcolor{orange!25}0.0439 & 25.25 & \cellcolor{orange!25}0.9496 & \cellcolor{orange!25}0.0479 \\
\textbf{Ours} (MSE) & \cellcolor{orange!25}26.55 & \cellcolor{red!25}0.9517 & 0.0518 & \cellcolor{orange!25}25.73 & \cellcolor{red!25}0.9523 & 0.0543 
\end{tabular}

\end{table}

\begin{figure}[tb]
  \centering
  \includegraphics[width=12.5cm]{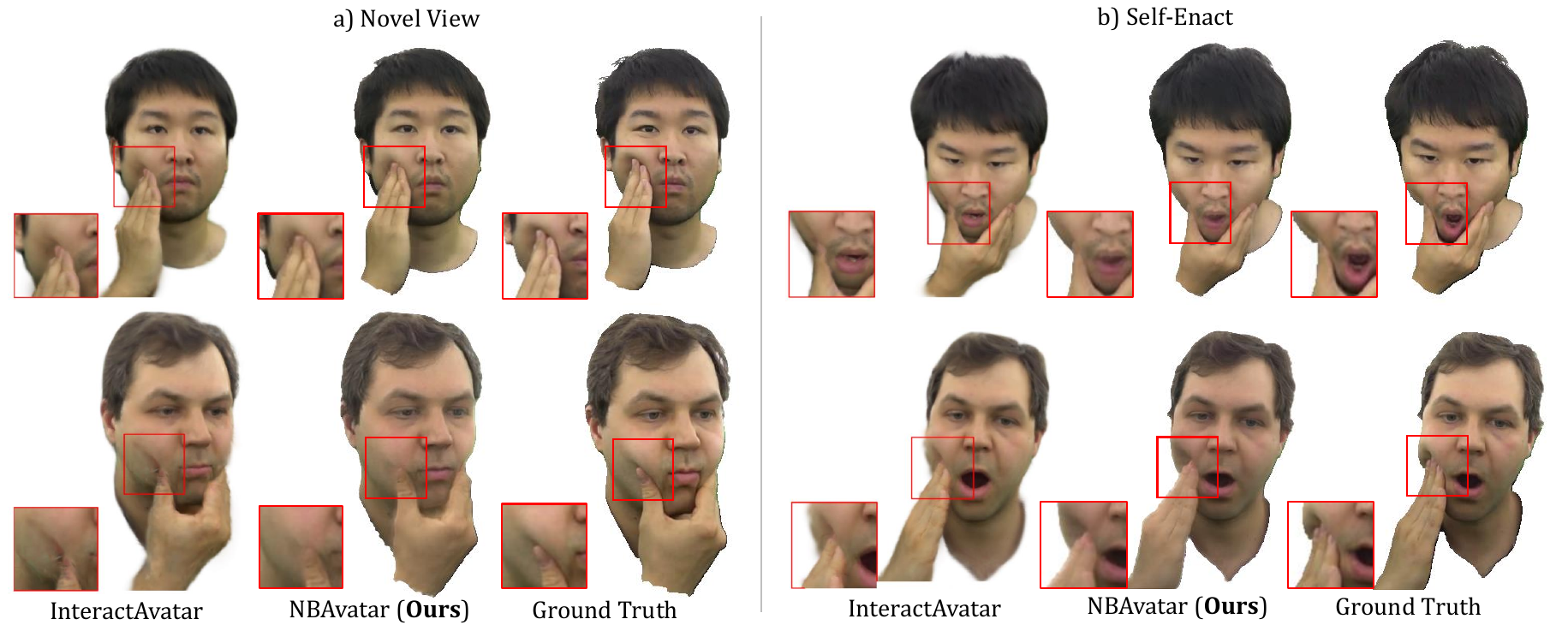}
  \caption{\textbf{Qualitative comparison with InteractAvatar~\cite{chen2025interactavatar}.} NBAvatar produces sharper details and more realistic hand-face deformations with fewer artifacts. InteractAvatar exhibits characteristic 3DGS artifacts including blurry facial textures and protruding Gaussians along the avatar boundary.}
  \label{fig:qualitative_ia}
\end{figure}

\subsection{Quantitative Results}
\label{sec:quant_results}

In \Cref{tab:metrics}, we report the quantitative results for novel-view and self-enactment rendering of the avatars. For self-enactment, the avatars are driven by held-out poses from the same subject to evaluate under novel pose scenarios. We compared with state-of-the-art head reconstruction methods SplattingAvatar~\cite{shao2024splattingavatar} and GaussianAvatar~\cite{qian2024gaussianavatars} under the same evaluation protocol by using the same data preprocessing and retraining these methods on the same data. 

Our NBAvatar achieves the best average scores across all three metrics in direct comparison with SplattingAvatar and GaussianAvatar. Notably, NBAvatar reduces the mean perceptual distance (LPIPS) by 42\% relative to GaussianAvatars (0.0439 vs.\ 0.0756) and by 53\% relative to SplattingAvatar (0.0439 vs.\ 0.0942) for novel views. For novel poses NBAvatar demonstrates similar improvement on LPIPS providing state-of-the-art value. The proposed method also improves PSNR and SSIM across both novel view and novel pose experiments. 

We additionally report metrics for our approach with the renderer finetuned with MSE loss after the avatar optimization finished. The use of a separate renderer allows for quick finetuning for different use cases by training the renderer $R$ only. MSE finetuning provides an additional growth in PSNR and SSIM by producing a smoother reconstruction, which comes at the cost of increased LPIPS. NBAvatar allows selecting a renderer during inference for the specific task while keeping the same billboard-based representation of the avatar.

%\textbf{Comparison with the InteractAvatar.}
In \Cref{tab:ia_protocol}, we compare against the metrics reported by InteractAvatar~\cite{chen2025interactavatar} in their paper by copying the provided values. Since the evaluation code and preprocessed data used in~\cite{chen2025interactavatar} are not publicly available, we reproduced their protocol as closely as possible based on the paper's description. NBAvatar achieves the highest SSIM on both novel-view synthesis and self-reenactment tasks, indicating improved structural consistency of avatars. But we have to note that differences in the implementation of preprocessing and evaluation scripts also affect metrics. We therefore encourage the reader to primarily rely on  \Cref{tab:metrics}, where all methods are evaluated under identical, fully controlled conditions.

%Crucially, the qualitative comparison in \Cref{fig:qualitative_ia} demonstrates that NBAvatar produces visibly sharper facial details, more realistic contact deformations, and cleaner silhouettes than InteractAvatar, confirming that the quantitative gap under their protocol does not reflect the actual perceived rendering quality.

%------------------------------------------------------------------
\subsection{Qualitative Results}
\label{sec:qual_results}

\begin{figure}[tb]
  \centering
  \includegraphics[width=12.5cm]{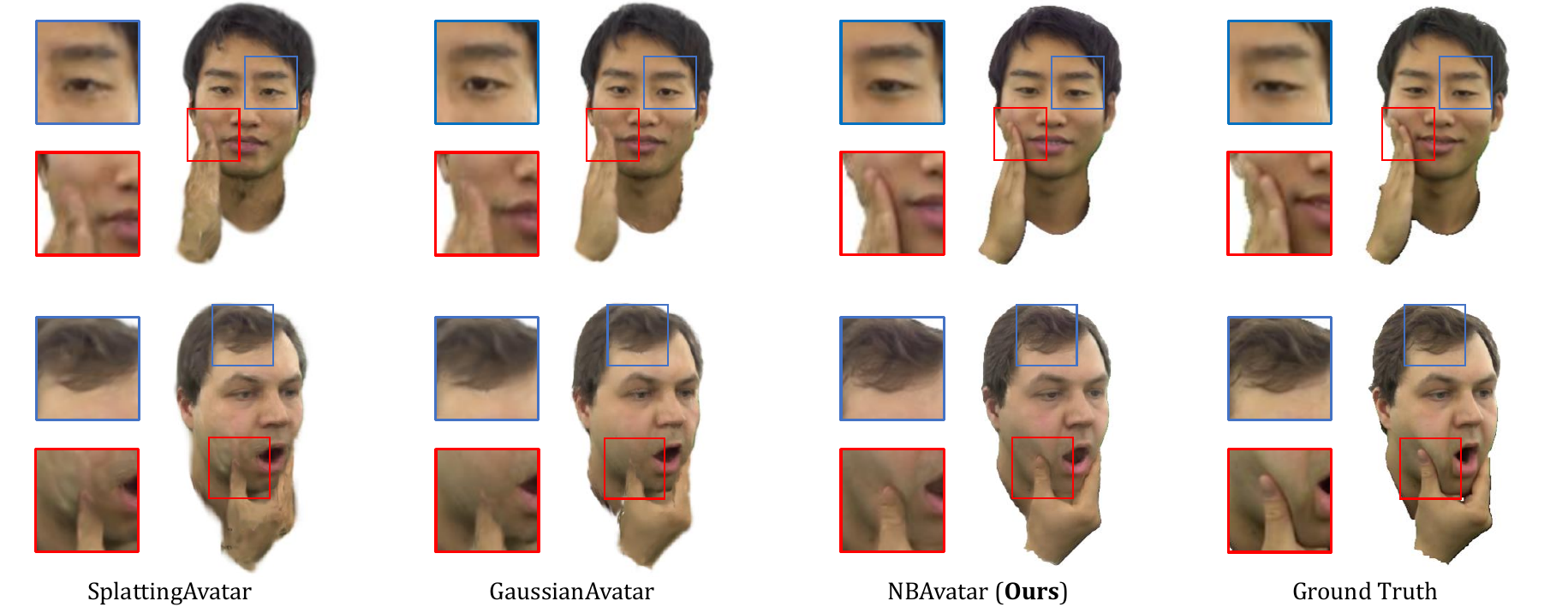}
  \caption{\textbf{Qualitative comparison on novel views.} 
  We compare our method with SplattingAvatar~\cite{shao2024splattingavatar} and GaussianAvatars~\cite{qian2024gaussianavatars} on the Decaf~\cite{shimada2023decaf} dataset. NBAvatar produces sharp, high-fidelity reconstructions of non-rigid facial deformations and dynamic hand appearance, accurately capturing facial details.
  %We compare our method (Ours), with SplattingAvatar~\cite{shao2024splattingavatar} (SA), and GaussianAvatars~\cite{qian2024gaussianavatars} (GA) on the Decaf~\cite{shimada2023decaf} dataset. NBAvatar produces sharp, high-fidelity reconstructions of non-rigid facial deformations and dynamic hand appearances, accurately capturing fine details such as shadows, wrinkles, and natural skin deformation in the contact regions.
  }
  \label{fig:qualitative}
\end{figure}

In \Cref{fig:qualitative_ia}, we qualitatively compared our results against those reported in the InteractAvatar~\cite{chen2025interactavatar} paper for novel views and poses scenarios. Our NBAvatar produces fewer artifacts, especially in border regions, and yields sharper facial details with increased structural similarity. The use of a hybrid implicit-explicit representation, where the renderer is combined with billboards, helps us reduce Gaussian-related artifacts and produce more realistic results than the baseline.

\Cref{fig:qualitative} presents a qualitative comparison on held-out novel views from the Decaf~\cite{shimada2023decaf} dataset with state-of-the-art face reconstruction methods. Both GaussianAvatars and SplattingAvatar demonstrate Gaussian-related artifacts such as overblurred regions and misplaced primitives caused by dynamic hand-face interaction. They are also subject to noticeable boundary artifacts where individual Gaussians protrude beyond the body silhouette. 
%In contrast, NBAvatar produces sharp, high-fidelity reconstructions that closely match the ground truth by producing faithful shadows in hand-face interaction regions due to spatial feature aggregation in the renderer $R$. The combination of explicit billboard geometry with implicit neural textures enables our method to accurately capture both the geometric deformations and the appearance dynamics of hand-face interaction. 
% From the rebuttal
In contrast, NBAvatar produces sharp, high-fidelity reconstructions by jointly leveraging both components of its hybrid representation. Explicit textured billboards preserve local details without Gaussian smoothing, while spatial feature aggregation in the neural renderer reconstructs high-frequency details and pose-dependent appearance.
Our Neural Billboard primitives, regularized to closely follow the mesh surface through our disentanglement training (\Cref{seq:disentanglement}), produce clean silhouettes free of the boundary artifacts common in Gaussian-based methods.

\begin{figure}[tb]
  \centering
  \includegraphics[width=12.5cm]{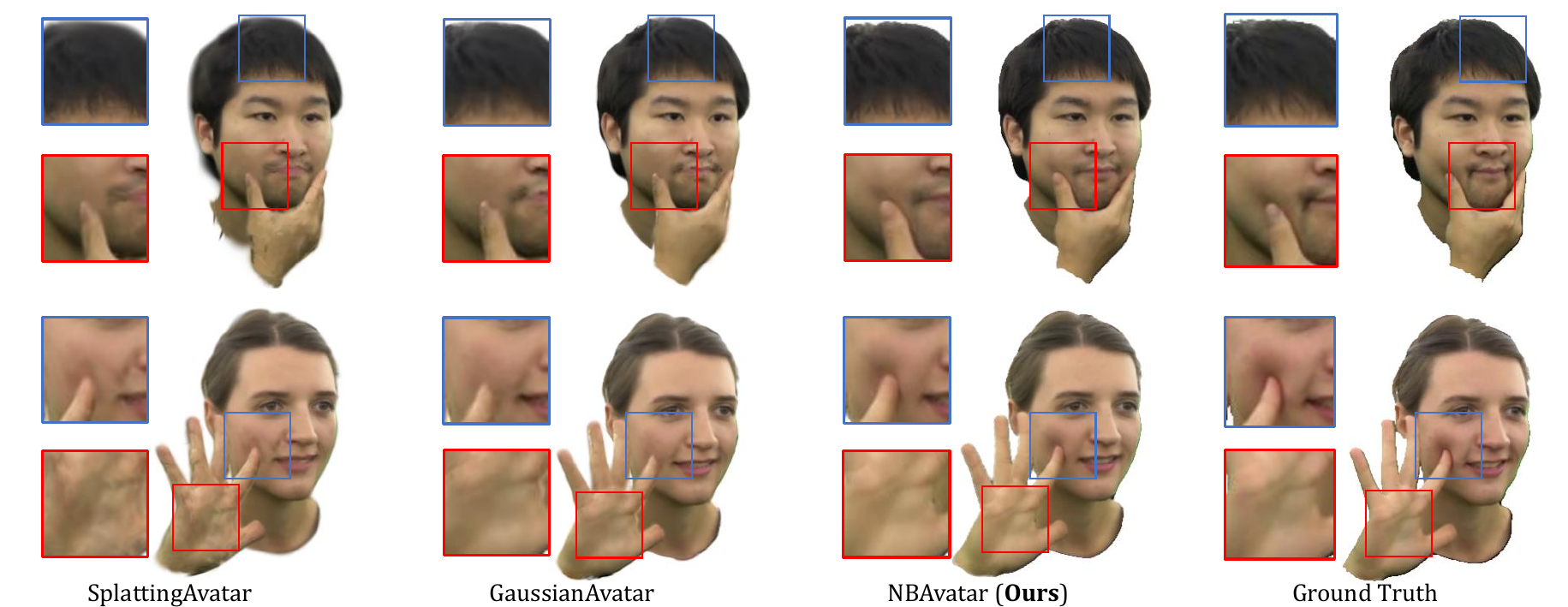}
  \caption{\textbf{Novel-pose synthesis (self-reenactment).} 
  We compare our method with SplattingAvatar~\cite{shao2024splattingavatar} and GaussianAvatars~\cite{qian2024gaussianavatars} on held-out poses from the Decaf~\cite{shimada2023decaf} dataset. NBAvatar generalizes to unseen hand-face interaction poses, faithfully reproducing contact-induced deformations and appearance changes.}
  \label{fig:qualitative_poses}
\end{figure}

We further evaluate the generalization of our method to unseen poses in \Cref{fig:qualitative_poses}. Novel-pose synthesis is particularly challenging for hand-face interaction avatars, as the model must correctly handle % deformations and 
appearance changes for contact configurations not seen during training. 
Both GaussianAvatars and SplattingAvatar struggle to adapt to novel poses, especially for rendering highly-articulated hands, resulting in an increasing number of artifacts.  
%Both GaussianAvatars and SplattingAvatar show degraded quality under novel poses: facial details become increasingly blurry, and the hand-face contact regions often exhibit implausible geometry or missing shadows. In contrast, 
NBAvatar maintains sharp, realistic renderings across unseen poses, demonstrating good generalization of the proposed Neural Billboard primitives. 
%The deferred neural rendering stage generalizes well to novel feature configurations, as the 
This is achieved by the UNet renderer~$R$ learning to interpret the spatial relationships between hand and face features rather than memorizing specific training poses. %This generalization ability is a direct benefit of our hybrid explicit-implicit representation, where the billboard geometry provides pose-consistent structure while the neural textures and renderer handle the appearance dynamics.

%------------------------------------------------------------------
\subsection{Ablation Study}
\label{sec:ablation}

\begin{table}[tb]
    \centering
    \caption{\textbf{Ablation study} on novel view synthesis for Decaf dataset.}
    \label{tab:ablation}
    \begin{tabular}{l c c c}
        \toprule
        Configuration & PSNR$\uparrow$ & SSIM$\uparrow$ & LPIPS$\downarrow$ \\
        \midrule
        %Full model & \textbf{TODO} & \textbf{TODO} & \textbf{TODO} \\
        %\midrule
        w/o $\mathcal{L}_{NB}$ & 25.41 & 0.9461 & 0.0453 \\
        w/o normals & 25.86 & 0.9474 & 0.0451 \\
        w/o poses & 25.12 & 0.9425 & 0.0509 \\
        w/o DNR &  24.34 & 0.9333 & 0.0655 \\
        Full Method & \cellcolor{red!25}26.08 & \cellcolor{red!25}0.9484 & \cellcolor{red!25}0.0439 \\
        %\bottomrule
    \end{tabular}
\end{table}

\begin{figure}[tb]
  \centering
  \includegraphics[width=12.2cm]{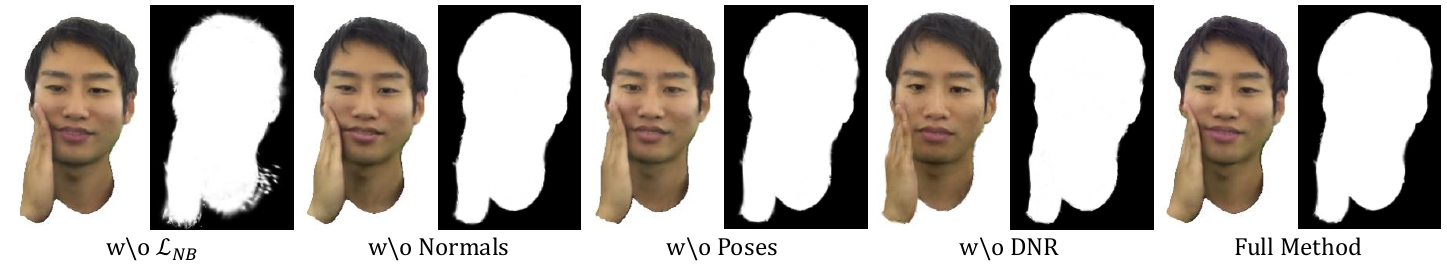}
  \caption{\textbf{Ablation study.} We demonstrate renders and billboards' masks for one of the Decaf subjects with ablated key components of the proposed algorithm.}
  \label{fig:ablation}
\end{figure}

We conducted an ablation study of the key components of the proposed approach in \Cref{tab:ablation} and showcase corresponding examples in \Cref{fig:ablation}. Namely, we ablated the proposed segmentation $\mathcal{L}_{NB}$ for disentanglement geometry and appearance. The exclusion of this loss results in noisy billboard orientations, leading to less consistent rendering and a significant reduction in objective metrics. We also ablated the use of normal maps for renderer $R$, demonstrating that while they provide an additional increase in metrics, the renderer $R$ is robust to working with features only. Next, we ablate the use of AdaIN-based pose broadcasting within the renderer feature maps. While this has a minor effect on rendering quality, it improves objective metrics by helping the network compensate for imperfect FLAME/MANO fits, leading to better matching with the ground truth. Finally, we ablated the use of deferred rendering (DNR) and demonstrated that it is essential for handling hand-caused shadowing and non-rigid deformations. We conclude that all parts of the proposed approach are essential to get the best qualitative result and objective metrics.

\begin{figure}[tb]
  \centering
  \includegraphics[width=12.2cm]{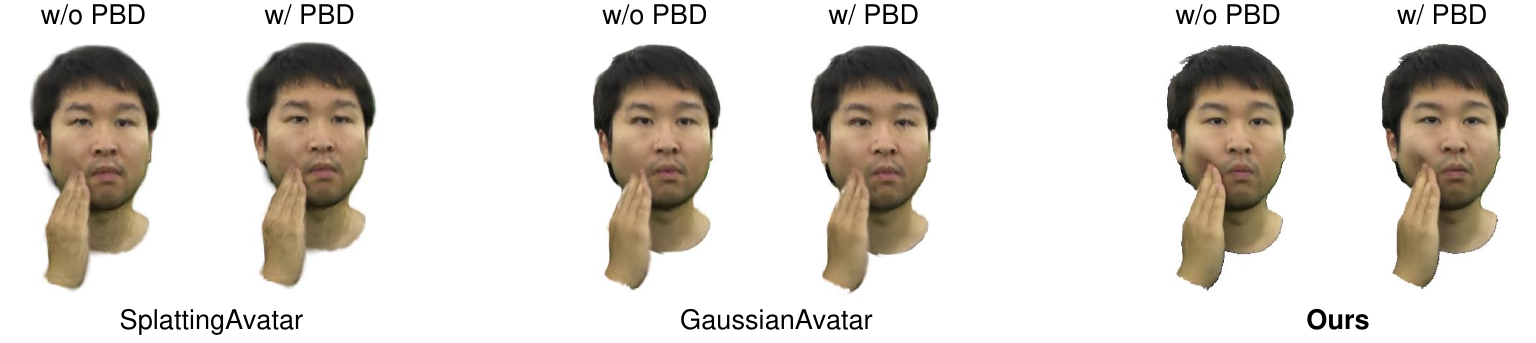}
  \caption{\textbf{Position-based dynamics ablation.} We report examples of rendering all methods under novel poses with PBD and without.}
  \label{fig:ablation_pbd}
\end{figure}

We also ablate the use of position-based dynamics (PBD)~\cite{muller2007position} in our method and present experiments on extending the baselines by introducing PBD. We showcase the rendering examples in the \Cref{fig:ablation_pbd} and report objective metrics in the \Cref{tab:ablation_bpd}. We find that while PBD provides slight metric improvements over the baselines, our approach surpasses them both with and without PBD. It is enabled by our renderer $R$ handling appearance changes without explicitly modeling geometry deformation.

\begin{table}[tb]
    \centering
    \caption{\textbf{Position-based dynamics ablation} for all methods under novel views.}
    \label{tab:ablation_bpd}
    \setlength{\tabcolsep}{5pt}
    \begin{tabular}{l|ccc|ccc}
                    & \multicolumn{3}{c|}{w/ PBD} & \multicolumn{3}{c}{w/o PBD} \\ \hline
                    & PSNR$\uparrow$ & SSIM$\uparrow$ & LPIPS$\downarrow$ & PSNR$\uparrow$ & SSIM$\uparrow$ & LPIPS$\downarrow$ \\ \hline
    SplattingAvatar & 24.67 & 0.9313 & 0.0942 & 24.63 & 0.9312 & 0.0943 \\
    GaussianAvatar  & 24.95 & 0.9397 & 0.0756 & 24.86 & 0.9394 & 0.0755  \\
    \textbf{Ours}   & \cellcolor{red!25}26.08 & \cellcolor{red!25}0.9484 & \cellcolor{red!25}0.0439 & \cellcolor{red!25}25.84 & \cellcolor{red!25}0.9475 & \cellcolor{red!25}0.0455 
    \end{tabular}
\end{table}

\section{Conclusion}

In this paper, we presented NBAvatar, a method for photorealistic rendering of head avatars with dynamic hand-face interaction. We introduced Neural Billboards, a hybrid representation that combines surface-aligned planar primitives with deferred neural rendering, enabling flexible geometry together with high-frequency appearance synthesis. We further proposed a geometry-aware training scheme with intermediate silhouette supervision that stabilizes the joint optimization of billboard geometry and neural texture features. Finally, we showed that spatial feature aggregation in screen space can implicitly capture complex interaction effects without explicit interaction-conditioned modules. Experiments on the Decaf dataset demonstrate improved perceptual quality and objective metrics such as PSNR, SSIM, and LPIPS compared to Gaussian-based avatar methods. In particular, we demonstrated LPIPS reduction up to 53\% compared to the baselines.

\bibliographystyle{splncs04}
\bibliography{main}

@String(CVPR  = {IEEE Conf. Comput. Vis. Pattern Recog.})

@String(ICCV  = {Int. Conf. Comput. Vis.})

@String(ECCV  = {Eur. Conf. Comput. Vis.})

@String(TOG   = {ACM Trans. Graph.})

@String(CGF   = {Comput. Graph. Forum})

@String(CVPR  = {CVPR})

@String(ICCV  = {ICCV})

@String(ECCV  = {ECCV})

@String(TOG   = {ACM TOG})

@article{shimada2023decaf,
  title={Decaf: Monocular deformation capture for face and hand interactions},
  author={Shimada, Soshi and Golyanik, Vladislav and P{\'e}rez, Patrick and Theobalt, Christian},
  journal = TOG,
  volume={42},
  number={6},
  pages={1--16},
  year={2023},
  publisher={ACM New York, NY, USA}
}

@inproceedings{chen2025interactavatar,
  title={InteractAvatar: Modeling Hand-Face Interaction in Photorealistic Avatars with Deformable Gaussians},
  author={Chen, Kefan and Mohan, Sreyas and Theiss, Justin and Oprea, Sergiu and Sridhar, Srinath and Prakash, Aayush},
  booktitle = ICCV,
  pages={10410--10420},
  year={2025}
}

@inproceedings{he2025capturing,
  title={Capturing head avatar with hand contacts from a monocular video},
  author={He, Haonan and Zheng, Yufeng and Song, Jie},
  booktitle = ICCV,
  pages={13099--13108},
  year={2025}
}

@inproceedings{wagner2025nephim,
  title={NePHIM: A Neural Physics-Based Head-Hand Interaction Model},
  author={Wagner, Nicolas and Schwanecke, Ulrich and Botsch, Mario},
  booktitle = CGF,
  pages={e70045},
  year={2025},
  organization={Wiley Online Library}
}

@article{muller2007position,
  title={Position based dynamics},
  author={M{\"u}ller, Matthias and Heidelberger, Bruno and Hennix, Marcus and Ratcliff, John},
  journal={Journal of Visual Communication and Image Representation},
  volume={18},
  number={2},
  pages={109--118},
  year={2007},
  publisher={Elsevier}
}

@article{thies2019deferred,
  title={Deferred neural rendering: Image synthesis using neural textures},
  author={Thies, Justus and Zollh{\"o}fer, Michael and Nie{\ss}ner, Matthias},
  journal = TOG,
  volume={38},
  number={4},
  pages={1--12},
  year={2019},
  publisher={ACM New York, NY, USA}
}

@inproceedings{svitov2025billboard,
  title={Billboard splatting (bbsplat): Learnable textured primitives for novel view synthesis},
  author={Svitov, David and Morerio, Pietro and Agapito, Lourdes and Del Bue, Alessio},
  booktitle={ICCV},
  pages={25029--25039},
  year={2025}
}

@article{kerbl20233d,
  title={3D Gaussian splatting for real-time radiance field rendering.},
  author={Kerbl, Bernhard and Kopanas, Georgios and Leimk{\"u}hler, Thomas and Drettakis, George},
  journal = TOG,
  volume={42},
  number={4},
  pages={139--1},
  year={2023}
}

@article{dimond1984face,
  title={Face touching in monkeys, apes and man: Evolutionary origins and cerebral asymmetry},
  author={Dimond, Stuart and Harries, Rashida},
  journal={Neuropsychologia},
  volume={22},
  number={2},
  pages={227--233},
  year={1984},
  publisher={Elsevier}
}

@article{mueller2019self,
  title={Self-touch: contact durations and point of touch of spontaneous facial self-touches differ depending on cognitive and emotional load},
  author={Mueller, Stephanie Margarete and Martin, Sven and Grunwald, Martin},
  journal={PloS one},
  volume={14},
  number={3},
  pages={e0213677},
  year={2019},
  publisher={Public Library of Science San Francisco, CA USA}
}

@inproceedings{shao2024splattingavatar,
  title={Splattingavatar: Realistic real-time human avatars with mesh-embedded gaussian splatting},
  author={Shao, Zhijing and Wang, Zhaolong and Li, Zhuang and Wang, Duotun and Lin, Xiangru and Zhang, Yu and Fan, Mingming and Wang, Zeyu},
  booktitle = CVPR,
  pages={1606--1616},
  year={2024}
}

@inproceedings{qian2024gaussianavatars,
  title={Gaussianavatars: Photorealistic head avatars with rigged 3d gaussians},
  author={Qian, Shenhan and Kirschstein, Tobias and Schoneveld, Liam and Davoli, Davide and Giebenhain, Simon and Nie{\ss}ner, Matthias},
  booktitle = CVPR,
  pages={20299--20309},
  year={2024}
}

@inproceedings{saito2024relightable,
  title={Relightable gaussian codec avatars},
  author={Saito, Shunsuke and Schwartz, Gabriel and Simon, Tomas and Li, Junxuan and Nam, Giljoo},
  booktitle = CVPR,
  pages={130--141},
  year={2024}
}

@inproceedings{li2024uravatar,
  title={Uravatar: Universal relightable gaussian codec avatars},
  author={Li, Junxuan and Cao, Chen and Schwartz, Gabriel and Khirodkar, Rawal and Richardt, Christian and Simon, Tomas and Sheikh, Yaser and Saito, Shunsuke},
  booktitle={SIGGRAPH Asia 2024 Conference Papers},
  pages={1--11},
  year={2024}
}

@inproceedings{grassal2022neural,
  title={Neural head avatars from monocular rgb videos},
  author={Grassal, Philip-William and Prinzler, Malte and Leistner, Titus and Rother, Carsten and Nie{\ss}ner, Matthias and Thies, Justus},
  booktitle = CVPR,
  pages={18653--18664},
  year={2022}
}

@inproceedings{zielonka2023instant,
  title={Instant volumetric head avatars},
  author={Zielonka, Wojciech and Bolkart, Timo and Thies, Justus},
  booktitle = CVPR,
  pages={4574--4584},
  year={2023}
}

@inproceedings{zheng2023pointavatar,
  title={Pointavatar: Deformable point-based head avatars from videos},
  author={Zheng, Yufeng and Yifan, Wang and Wetzstein, Gordon and Black, Michael J and Hilliges, Otmar},
  booktitle = CVPR,
  pages={21057--21067},
  year={2023}
}

@inproceedings{gafni2021dynamic,
  title={Dynamic neural radiance fields for monocular 4d facial avatar reconstruction},
  author={Gafni, Guy and Thies, Justus and Zollhofer, Michael and Nie{\ss}ner, Matthias},
  booktitle = CVPR,
  pages={8649--8658},
  year={2021}
}

@inproceedings{xu2023avatarmav,
  title={Avatarmav: Fast 3d head avatar reconstruction using motion-aware neural voxels},
  author={Xu, Yuelang and Wang, Lizhen and Zhao, Xiaochen and Zhang, Hongwen and Liu, Yebin},
  booktitle={ACM SIGGRAPH 2023 Conference Proceedings},
  pages={1--10},
  year={2023}
}

@article{svitov2025closeupavatar,
  title={CloseUpAvatar: High-Fidelity Animatable Full-Body Avatars with Mixture of Multi-Scale Textures},
  author={Svitov, David and Morerio, Pietro and Agapito, Lourdes and Del Bue, Alessio},
  journal={arXiv preprint arXiv:2512.03593},
  year={2025}
}

@inproceedings{zheng2022avatar,
  title={Im avatar: Implicit morphable head avatars from videos},
  author={Zheng, Yufeng and Abrevaya, Victoria Fern{\'a}ndez and B{\"u}hler, Marcel C and Chen, Xu and Black, Michael J and Hilliges, Otmar},
  booktitle = CVPR,
  pages={13545--13555},
  year={2022}
}

@article{muller2022instant,
  title={Instant neural graphics primitives with a multiresolution hash encoding},
  author={M{\"u}ller, Thomas and Evans, Alex and Schied, Christoph and Keller, Alexander},
  journal = TOG,
  volume={41},
  number={4},
  pages={1--15},
  year={2022},
  publisher={ACM New York, NY, USA}
}

@article{li2017learning,
  title={Learning a model of facial shape and expression from 4D scans.},
  author={Li, Tianye and Bolkart, Timo and Black, Michael J and Li, Hao and Romero, Javier},
  journal = TOG,
  volume={36},
  number={6},
  pages={194--1},
  year={2017}
}

@inproceedings{buhler2021varitex,
  title={Varitex: Variational neural face textures},
  author={B{\"u}hler, Marcel C and Meka, Abhimitra and Li, Gengyan and Beeler, Thabo and Hilliges, Otmar},
  booktitle = ICCV,
  pages={13890--13899},
  year={2021}
}

@article{romero2022embodied,
  title={Embodied hands: Modeling and capturing hands and bodies together},
  author={Romero, Javier and Tzionas, Dimitrios and Black, Michael J},
  journal={arXiv preprint arXiv:2201.02610},
  year={2022}
}

@inproceedings{mildenhall2020nerf,
 title={NeRF: Representing Scenes as Neural Radiance Fields for View Synthesis},
 author={Ben Mildenhall and Pratul P. Srinivasan and Matthew Tancik and Jonathan T. Barron and Ravi Ramamoorthi and Ren Ng},
 year={2020},
 booktitle= ECCV,
}

@article{feng2021learning,
  title={Learning an animatable detailed 3D face model from in-the-wild images},
  author={Feng, Yao and Feng, Haiwen and Black, Michael J and Bolkart, Timo},
  journal= TOG,
  volume={40},
  number={4},
  pages={1--13},
  year={2021},
  publisher={ACM New York, NY, USA}
}

@article{lombardi2018deep,
  title={Deep appearance models for face rendering},
  author={Lombardi, Stephen and Saragih, Jason and Simon, Tomas and Sheikh, Yaser},
  journal = TOG,
  volume={37},
  number={4},
  pages={1--13},
  year={2018},
  publisher={ACM New York, NY, USA}
}

@incollection{blanz2023morphable,
  title={A morphable model for the synthesis of 3D faces},
  author={Blanz, Volker and Vetter, Thomas},
  booktitle={Seminal Graphics Papers: Pushing the Boundaries, Volume 2},
  pages={157--164},
  year={2023}
}

@inproceedings{blanz2003reanimating,
  title={Reanimating faces in images and video},
  author={Blanz, Volker and Basso, Curzio and Poggio, Tomaso and Vetter, Thomas},
  booktitle = CGF,
  volume={22},
  number={3},
  pages={641--650},
  year={2003},
  organization={Wiley Online Library}
}

@inproceedings{paysan20093d,
  title={A 3D face model for pose and illumination invariant face recognition},
  author={Paysan, Pascal and Knothe, Reinhard and Amberg, Brian and Romdhani, Sami and Vetter, Thomas},
  booktitle={2009 sixth IEEE international conference on advanced video and signal based surveillance},
  pages={296--301},
  year={2009},
  organization={Ieee}
}

@inproceedings{gecer2019ganfit,
  title={Ganfit: Generative adversarial network fitting for high fidelity 3d face reconstruction},
  author={Gecer, Baris and Ploumpis, Stylianos and Kotsia, Irene and Zafeiriou, Stefanos},
  booktitle = CVPR,
  pages={1155--1164},
  year={2019}
}

@inproceedings{huang20242d,
  title={2d gaussian splatting for geometrically accurate radiance fields},
  author={Huang, Binbin and Yu, Zehao and Chen, Anpei and Geiger, Andreas and Gao, Shenghua},
  booktitle={ACM SIGGRAPH 2024 conference papers},
  pages={1--11},
  year={2024}
}

@inproceedings{corona2022lisa,
  title={Lisa: Learning implicit shape and appearance of hands},
  author={Corona, Enric and Hodan, Tomas and Vo, Minh and Moreno-Noguer, Francesc and Sweeney, Chris and Newcombe, Richard and Ma, Lingni},
  booktitle = CVPR,
  pages={20533--20543},
  year={2022}
}

@inproceedings{mundra2023livehand,
  title={Livehand: Real-time and photorealistic neural hand rendering},
  author={Mundra, Akshay and Wang, Jiayi and Habermann, Marc and Theobalt, Christian and Elgharib, Mohamed and others},
  booktitle = ICCV,
  pages={18035--18045},
  year={2023}
}

@inproceedings{pokhariya2024manus,
  title={Manus: Markerless grasp capture using articulated 3d gaussians},
  author={Pokhariya, Chandradeep and Shah, Ishaan Nikhil and Xing, Angela and Li, Zekun and Chen, Kefan and Sharma, Avinash and Sridhar, Srinath},
  booktitle = CVPR,
  pages={2197--2208},
  year={2024}
}

@article{wu2024dice,
  title={Dice: End-to-end deformation capture of hand-face interactions from a single image},
  author={Wu, Qingxuan and Dou, Zhiyang and Xu, Sirui and Shimada, Soshi and Wang, Chen and Yu, Zhengming and Liu, Yuan and Lin, Cheng and Cao, Zeyu and Komura, Taku and others},
  journal={arXiv preprint arXiv:2406.17988},
  year={2024}
}

@article{kirschstein2023nersemble,
  title={Nersemble: Multi-view radiance field reconstruction of human heads},
  author={Kirschstein, Tobias and Qian, Shenhan and Giebenhain, Simon and Walter, Tim and Nie{\ss}ner, Matthias},
  journal = TOG,
  volume={42},
  number={4},
  pages={1--14},
  year={2023},
  publisher={ACM New York, NY, USA}
}

@article{weiss2024gaussian,
  title={Gaussian billboards: Expressive 2d gaussian splatting with textures},
  author={Weiss, Sebastian and Bradley, Derek},
  journal={arXiv preprint arXiv:2412.12734},
  year={2024}
}

@inproceedings{rong2025gstex,
  title={Gstex: Per-primitive texturing of 2d gaussian splatting for decoupled appearance and geometry modeling},
  author={Rong, Victor and Chen, Jingxiang and Bahmani, Sherwin and Kutulakos, Kiriakos N and Lindell, David B},
  booktitle={2025 IEEE/CVF Winter Conference on Applications of Computer Vision (WACV)},
  pages={3508--3518},
  year={2025},
  organization={IEEE}
}

@article{song2024hdgs,
  title={Hdgs: Textured 2d gaussian splatting for enhanced scene rendering},
  author={Song, Yunzhou and Lin, Heguang and Lei, Jiahui and Liu, Lingjie and Daniilidis, Kostas},
  journal={arXiv preprint arXiv:2412.01823},
  year={2024}
}

@article{xu2024supergaussians,
  title={SuperGaussians: Enhancing Gaussian Splatting Using Primitives with Spatially Varying Colors},
  author={Xu, Rui and Chen, Wenyue and Wang, Jiepeng and Liu, Yuan and Wang, Peng and Gao, Lin and Xin, Shiqing and Komura, Taku and Li, Xin and Wang, Wenping},
  journal={arXiv preprint arXiv:2411.18966},
  year={2024}
}

@inproceedings{sigg2006gpu,
  title={GPU-based ray-casting of quadratic surfaces.},
  author={Sigg, Christian and Weyrich, Tim and Botsch, Mario and Gross, Markus H},
  booktitle={PBG@SIGGRAPH},
  pages={59--65},
  year={2006}
}

@inproceedings{svitov2024haha,
  title={Haha: Highly articulated gaussian human avatars with textured mesh prior},
  author={Svitov, David and Morerio, Pietro and Agapito, Lourdes and Del Bue, Alessio},
  booktitle={Proceedings of the Asian Conference on Computer Vision},
  pages={4051--4068},
  year={2024}
}

@article{carion2025sam,
  title={Sam 3: Segment anything with concepts},
  author={Carion, Nicolas and Gustafson, Laura and Hu, Yuan-Ting and Debnath, Shoubhik and Hu, Ronghang and Suris, Didac and Ryali, Chaitanya and Alwala, Kalyan Vasudev and Khedr, Haitham and Huang, Andrew and others},
  journal={arXiv preprint arXiv:2511.16719},
  year={2025}
}

@inproceedings{ronneberger2015u,
  title={U-net: Convolutional networks for biomedical image segmentation},
  author={Ronneberger, Olaf and Fischer, Philipp and Brox, Thomas},
  booktitle={International Conference on Medical image computing and computer-assisted intervention},
  pages={234--241},
  year={2015},
  organization={Springer}
}

@inproceedings{grigorev2021stylepeople,
  title={Stylepeople: A generative model of fullbody human avatars},
  author={Grigorev, Artur and Iskakov, Karim and Ianina, Anastasia and Bashirov, Renat and Zakharkin, Ilya and Vakhitov, Alexander and Lempitsky, Victor},
  booktitle = CVPR,
  pages={5151--5160},
  year={2021}
}

@inproceedings{bender2015position,
  title={Position-Based Simulation Methods in Computer Graphics.},
  author={Bender, Jan and M{\"u}ller, Matthias and Macklin, Miles},
  booktitle={Eurographics (tutorials)},
  pages={1--32},
  year={2015}
}

@inproceedings{chentanez2020cloth,
  title={Cloth and skin deformation with a triangle mesh based convolutional neural network},
  author={Chentanez, Nuttapong and Macklin, Miles and M{\"u}ller, Matthias and Jeschke, Stefan and Kim, Tae-Yong},
  booktitle = CGF,
  volume={39},
  number={8},
  pages={123--134},
  year={2020},
  organization={Wiley Online Library}
}

@article{sun2024physhand,
  title={Physhand: A hand simulation model with physiological geometry, physical deformation, and accurate contact handling},
  author={Sun, Mingyang and Kou, Dongliang and Yuan, Ruisheng and Yang, Dingkang and Zhai, Peng and Zhao, Xiao and Jiang, Yang and Li, Xiong and Li, Jingchen and Zhang, Lihua},
  journal={arXiv preprint arXiv:2409.05143},
  year={2024}
}

@inproceedings{milletari2016v,
  title={V-net: Fully convolutional neural networks for volumetric medical image segmentation},
  author={Milletari, Fausto and Navab, Nassir and Ahmadi, Seyed-Ahmad},
  booktitle={2016 fourth international conference on 3D vision (3DV)},
  pages={565--571},
  year={2016},
  organization={Ieee}
}

@inproceedings{zhang2018unreasonable,
  title={The unreasonable effectiveness of deep features as a perceptual metric},
  author={Zhang, Richard and Isola, Phillip and Efros, Alexei A and Shechtman, Eli and Wang, Oliver},
  booktitle = CVPR,
  pages={586--595},
  year={2018}
}

@inproceedings{bashirov2024morf,
  title={Morf: Mobile realistic fullbody avatars from a monocular video},
  author={Bashirov, Renat and Larionov, Alexey and Ustinova, Evgeniya and Sidorenko, Mikhail and Svitov, David and Zakharkin, Ilya and Lempitsky, Victor},
  booktitle={Proceedings of the IEEE/CVF Winter Conference on Applications of Computer Vision},
  pages={3545--3555},
  year={2024}
}

@article{qi2017pointnet++,
  title={Pointnet++: Deep hierarchical feature learning on point sets in a metric space},
  author={Qi, Charles Ruizhongtai and Yi, Li and Su, Hao and Guibas, Leonidas J},
  journal={Advances in neural information processing systems},
  volume={30},
  year={2017}
}

@article{radl2024stopthepop,
  title={Stopthepop: Sorted gaussian splatting for view-consistent real-time rendering},
  author={Radl, Lukas and Steiner, Michael and Parger, Mathias and Weinrauch, Alexander and Kerbl, Bernhard and Steinberger, Markus},
  journal = TOG,
  volume={43},
  number={4},
  pages={1--17},
  year={2024},
  publisher={ACM New York, NY, USA}
}

@article{kingma2014adam,
  title={Adam: A method for stochastic optimization},
  author={Kingma, Diederik P},
  journal={arXiv preprint arXiv:1412.6980},
  year={2014}
}

@inproceedings{chen2023hand,
  title={Hand avatar: Free-pose hand animation and rendering from monocular video},
  author={Chen, Xingyu and Wang, Baoyuan and Shum, Heung-Yeung},
  booktitle=CVPR,
  pages={8683--8693},
  year={2023}
}

@inproceedings{kalshetti2024intrinsic,
  title={Intrinsic hand avatar: Illumination-aware hand appearance and shape reconstruction from monocular rgb video},
  author={Kalshetti, Pratik and Chaudhuri, Parag},
  booktitle={Proceedings of the IEEE/CVF Winter Conference on Applications of Computer Vision},
  pages={6120--6130},
  year={2024}
}

@inproceedings{guo2023handnerf,
  title={Handnerf: Neural radiance fields for animatable interacting hands},
  author={Guo, Zhiyang and Zhou, Wengang and Wang, Min and Li, Li and Li, Houqiang},
  booktitle=CVPR,
  pages={21078--21087},
  year={2023}
}

@article{guo2025handnerf++,
  title={HandNeRF++: Modeling Animatable Interacting Hands With Neural Radiance Fields},
  author={Guo, Zhiyang and Zhou, Wengang and Wang, Min and Li, Li and Li, Houqiang},
  journal={IEEE Transactions on Pattern Analysis and Machine Intelligence},
  year={2025},
  publisher={IEEE}
}

@article{wang2004image,
  title={Image quality assessment: from error visibility to structural similarity},
  author={Wang, Zhou and Bovik, Alan C and Sheikh, Hamid R and Simoncelli, Eero P},
  journal={IEEE transactions on image processing},
  volume={13},
  number={4},
  pages={600--612},
  year={2004},
  publisher={IEEE}
}

@inproceedings{fridovich2022plenoxels,
  title={Plenoxels: Radiance fields without neural networks},
  author={Fridovich-Keil, Sara and Yu, Alex and Tancik, Matthew and Chen, Qinhong and Recht, Benjamin and Kanazawa, Angjoo},
  booktitle = CVPR,
  pages={5501--5510},
  year={2022}
}

@inproceedings{chen2022tensorf,
  title={Tensorf: Tensorial radiance fields},
  author={Chen, Anpei and Xu, Zexiang and Geiger, Andreas and Yu, Jingyi and Su, Hao},
  booktitle = ECCV,
  pages={333--350},
  year={2022},
  organization={Springer}
}

@inproceedings{huang2017arbitrary,
  title={Arbitrary style transfer in real-time with adaptive instance normalization},
  author={Huang, Xun and Belongie, Serge},
  booktitle= ICCV,
  pages={1501--1510},
  year={2017}
}

\clearpage
\setcounter{section}{0}
\renewcommand{\thesection}{\Alph{section}}

%\begin{comment}
% ---------------------------------------------------------------
 
%% redefine the \title command so that a variable name is saved in \thetitle, and provides the \maketitlesupplementary command 
\let\titleold\title
\renewcommand{\title}[1]{\titleold{#1}\newcommand{\thetitle}{#1}}
\def\maketitlesupplementary
   {
   %\newpage
   %    \twocolumn[
        %\centering
    %    \Large
    %    \textbf{\thetitle}\\
    %    \vspace{0.5em}Supplementary Material \\
    %    \vspace{1.0em}
   %    ] %< twocolumn
   \onecolumn
   \begin{center}
   \Large
    \textbf{\thetitle}\\
    \small
    \vspace{0.5em}------------------------------ \\
    \vspace{1.0em}
   \end{center}
    % \centering
    % \Large
    % \textbf{\thetitle}\\
    % \vspace{0.5em}Supplementary Material \\
    % \vspace{1.0em}
     %< twocolumn
   }
 
% ---------------------------------------------------------------

\title{Supplementary materials for "NBAvatar: Neural Billboards Avatars with Realistic \\ Hand-Face Interaction"}
\maketitlesupplementary
\suppressfloats[t]

\section{Implementation Details}

\subsection{Training Details}
\textbf{Optimization details.} All trainable parameters are optimized using the Adam optimizer with default momentum parameters $(\beta_1=0.9, \beta_2=0.999)$. The renderer network $R$ is trained with a learning rate of $1\times10^{-3}$. Billboard color parameters are optimized with a learning rate of $2.5\times10^{-3}$, while billboard rotations and scales use learning rates of $1\times10^{-3}$ and $5\times10^{-3}$, respectively. The neural texture parameters are optimized with learning rates of $5\times10^{-4}$ for both the feature texture $T_i^{NT}$ and the opacity texture $T_i^{\alpha}$. Billboard position offsets $\mu_i$ are optimized with a learning rate of $1.6\times10^{-4}$, and their learning rate is exponentially decayed during training with factor $\gamma = 0.01^{1/N}$, where $N=100,000$ and denotes the total number of optimization steps.

\textbf{Spectral initialization.}
To stabilize the optimization of neural billboard textures, we initialize the feature textures using spectral coordinates derived from the underlying mesh geometry. Specifically, we construct the cotangent Laplacian of the mesh and extract several of its lowest-frequency eigenvectors, which form a smooth spectral basis over the surface. Intuitively, these eigenvectors act as globally consistent coordinate functions that vary smoothly across the mesh geometry. We use these spectral coordinates as initial feature channels for the billboard textures, providing a structured low-frequency embedding of the surface that helps the renderer converge to meaningful spatial representations and gradually learn high-frequency appearance details.

\textbf{Train-Test camera split.} Following the evaluation protocol used for the Decaf dataset in InteractAvatar, we train the avatars using seven camera views and reserve one camera for novel-view evaluation. The exact train–test camera splits used for each subject are provided in \Cref{tab:novel_view_cameras}.

\begin{table}[tb]
    \centering
    \caption{\textbf{Cameras setup.} We used the following train-test camera split for the Decaf dataset for each subject.}
    \label{tab:novel_view_cameras}
    \begin{tabular}{l c c}
        \toprule
        Subject & Training Cameras & Evaluation Cameras \\
        \midrule
        S1 & 110, 100, 108, 084, 111, 121, 122 & 102 \\
        S2 & 122, 110, 102, 100, 084, 111, 121 & 108 \\
        S3 & 121, 108, 122, 110, 102, 111, 084 & 100 \\
        S4 & 108, 122, 110, 102, 100, 084, 121 & 111 \\
        %\bottomrule
    \end{tabular}
\end{table}

\textbf{Renderer architecture.} The neural renderer $R$ follows a U-Net encoder- decoder architecture with skip connections between corresponding resolution levels. The encoder consists of three downsampling stages implemented using stride-2 convolutions followed by AdaIn normalization and LeakyReLU activations. AdaIN layers condition the normalization on a pose embedding: the concatenated FLAME and MANO parameters are first encoded by a shared two-layer MLP into a fixed-size embedding, which is then linearly projected to produce per-channel affine parameters (scale and shift) applied after Instance Normalization. The decoder mirrors this structure and progressively reconstructs the image resolution. Instead of transposed convolutions, we employ bilinear upsampling followed by $3\times3$ convolutions to avoid checkerboard artifacts that commonly appear in deconvolution-based decoders. Skip connections concatenate encoder features with the corresponding decoder activations, allowing the renderer to preserve spatial details from the rasterized billboard feature map. The network predicts RGB color and opacity using two separate output heads followed by sigmoid activations.

\section{Inference Efficiency}

We evaluate the inference speed of the proposed pipeline by measuring the runtime of its main components, including billboard rasterization and neural rendering. As shown in \Cref{tab:inference_speed}, rasterization of billboard primitives is extremely efficient, taking only 1.3\,ms when neural textures are used. The main computational cost comes from the neural renderer, which converts the rasterized feature maps into the final RGB output. Using half-precision inference (fp16) significantly reduces the rendering time from 5.53\,ms to 2.97\,ms without affecting visual quality. Overall, the full pipeline runs in 4.13\,ms per frame, corresponding to 242 FPS. This demonstrates that our method achieves real-time performance. The renderer could be further accelerated using standard CNN optimization techniques such as pruning, knowledge distillation, or quantization.

\begin{table}[tb]
    \centering
    \caption{\textbf{Inference speed evaluation.} We report inference speed and corresponding FPS for different stages of the rendering pipeline. We additionally investigate the inference speed of rasterization with RGB and Neural Textures (NT).}
    \label{tab:inference_speed}
    \begin{tabular}{l c c}
        \toprule
        Stage & Speed (ms)$\downarrow$ & FPS$\uparrow$ \\
        \midrule
        Rasterization - RGB & 1.1 & 880 \\
        Rasterization - NT & 1.3 & 750 \\
        Neural Rendering - fp32 & 5.52 & 181 \\
        Neural Rendering - fp16 & 2.97 & 336 \\ \hline
        Total (NT; fp16) & 4.13 & 242 \\
        %\bottomrule
    \end{tabular}
\end{table}

\section{Per-Subject Results}

\begin{figure}[tb]
  \centering
  \includegraphics[width=12.0cm]{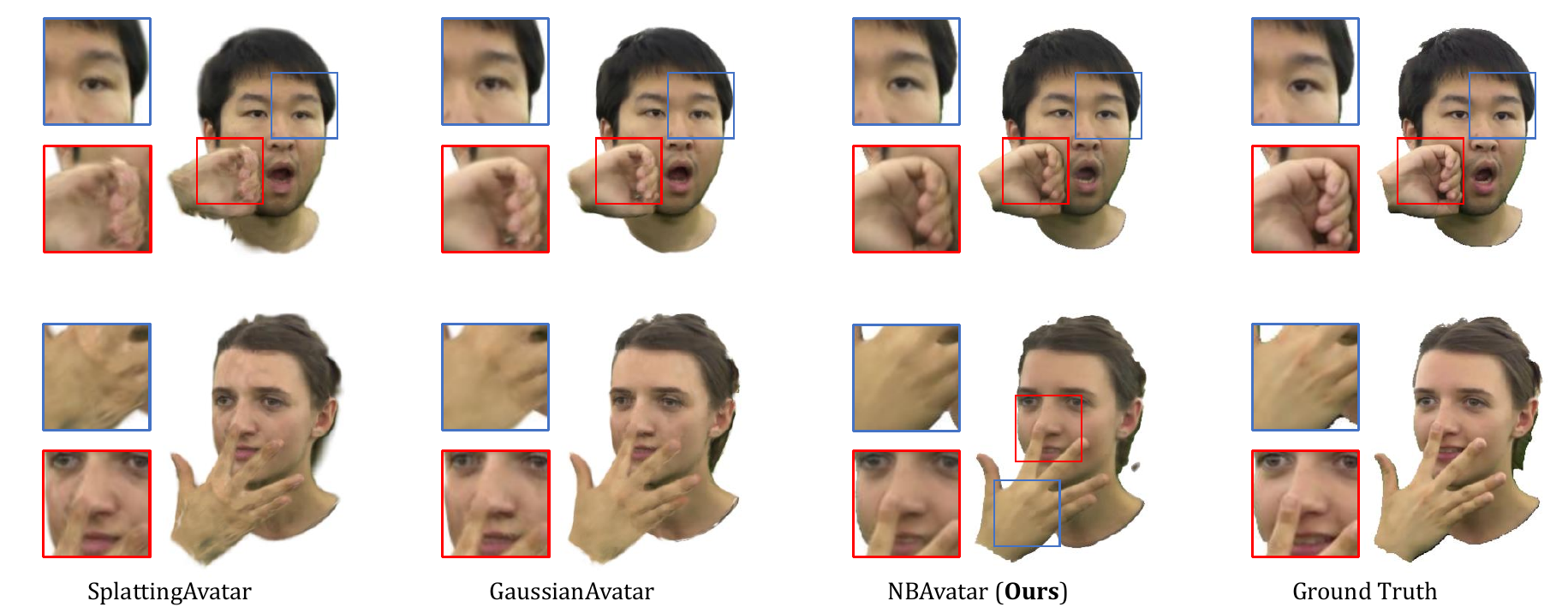}
  \caption{\textbf{Qualitative comparison on novel views for additional subjects.} 
  We compare our method with SplattingAvatar~\cite{shao2024splattingavatar} and GaussianAvatars~\cite{qian2024gaussianavatars} on the Decaf~\cite{shimada2023decaf} dataset. Our NBAvatar produces sharp, high-fidelity reconstructions of non-rigid facial deformations and dynamic hand appearance, accurately capturing facial details.}
  \label{fig:novel_views_supp_mat}
  \vspace{-10pt}
\end{figure}

\begin{figure}[tb]
  \centering
  \includegraphics[width=12.0cm]{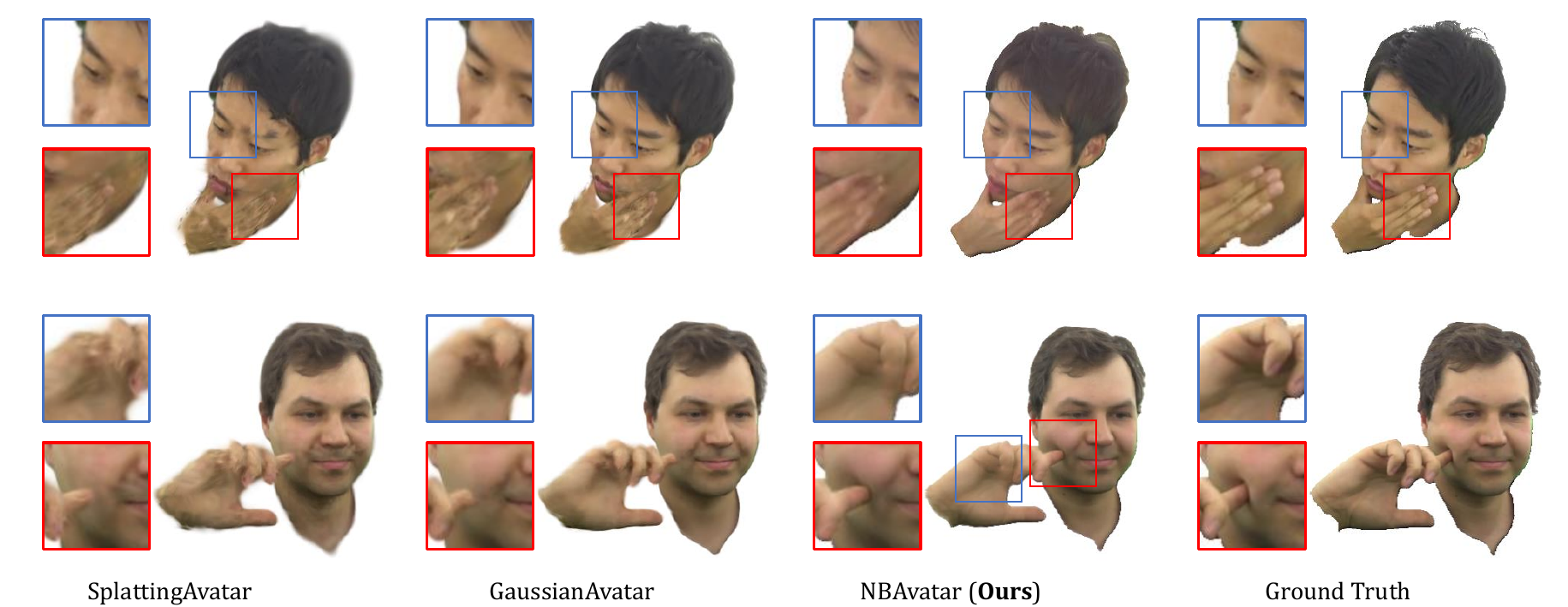}
  \caption{\textbf{Novel-pose synthesis for additional subjects.} 
  We compare our method with SplattingAvatar~\cite{shao2024splattingavatar} and GaussianAvatars~\cite{qian2024gaussianavatars} on held-out poses from the Decaf~\cite{shimada2023decaf} dataset. Our NBAvatar generalizes to unseen hand-face interaction poses, faithfully reproducing contact-induced deformations and appearance changes.}
  \label{fig:novel_poses_supp_mat}
  \vspace{-10pt}
\end{figure}

\begin{figure}[tb]
  \centering
  \includegraphics[width=12.0cm]{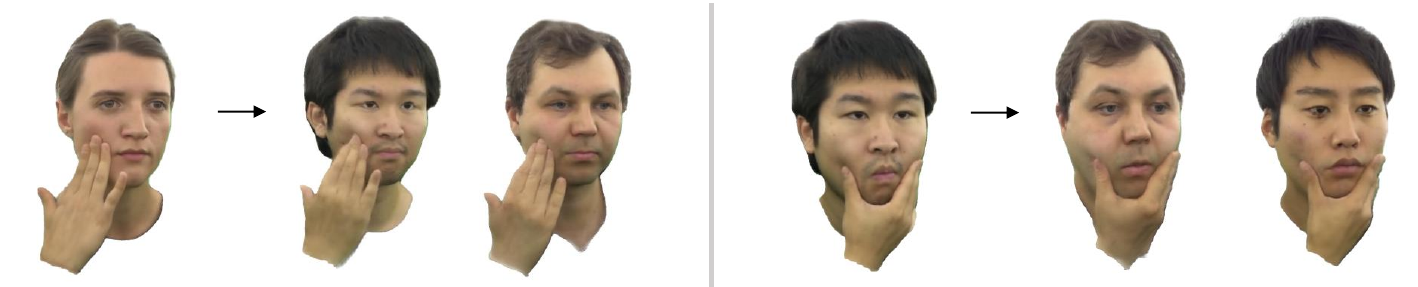}
  \caption{\textbf{Cross-reenactment for different subjects.} We report more examples of cross-reenactment when we drive one actor with the poses of another.}
  \label{fig:cross_actors}
  \vspace{-10pt}
\end{figure}

\textbf{Visual results.} In \Cref{fig:novel_views_supp_mat} and \Cref{fig:novel_poses_supp_mat}, we provide additional visual results for more subjects under novel view and poses scenarios. NBAvatar produces a more faithful reconstruction of hand–face interactions compared to baselines. In \Cref{fig:cross_actors} we also report more examples of cross-reenactment.

\textbf{Quantitative results per subject.} In \Cref{tab:novel_view_supp_matt} and \Cref{tab:novel_pose_supp_matt}, we report the quantitative results for each subject in the novel-view and novel-pose evaluations presented in the main paper. NBAvatar achieves the best PSNR, SSIM, and LPIPS for all subjects, except for subject S2 in the novel-pose evaluation. The performance drop on S2 is caused by the actor wearing a different T-shirt during the novel-pose recording. Although this appearance mismatch affects all methods, our approach is more sensitive to it because it more faithfully reconstructs the subject's appearance during training.

\begin{table}[t]
\centering
\caption{\textbf{Novel-view metrics per person.} We report novel-view quantitative objective metrics separately for each person used in the evaluation on the Decaf dataset.}
\label{tab:novel_view_supp_matt}
\resizebox{\textwidth}{!}{%
\begin{tabular}{l ccc ccc ccc ccc}
\toprule
& \multicolumn{3}{c}{S1} & \multicolumn{3}{c}{S2} & \multicolumn{3}{c}{S3} & \multicolumn{3}{c}{S4} \\
\cmidrule(lr){2-4} \cmidrule(lr){5-7} \cmidrule(lr){8-10} \cmidrule(lr){11-13}
Method & PSNR$\uparrow$ & SSIM$\uparrow$ & LPIPS$\downarrow$ & PSNR$\uparrow$ & SSIM$\uparrow$ & LPIPS$\downarrow$ & PSNR$\uparrow$ & SSIM$\uparrow$ & LPIPS$\downarrow$ & PSNR$\uparrow$ & SSIM$\uparrow$ & LPIPS$\downarrow$ \\
\midrule
Splatt. Av.~\cite{shao2024splattingavatar}
  & 25.17 & 0.9363 & 0.0892
  & 25.83 & 0.9501 & 0.0689
  & 24.45 & 0.9309 & 0.0938
  & 23.24 & 0.9080 & 0.1251 \\
Gauss. Av.~\cite{qian2024gaussianavatars}
  & 25.31 & 0.9436 & 0.0691
  & 26.55 & 0.9605 & 0.0533
  & 24.55 & 0.9390 & 0.0763
  & 23.40 & 0.9159 & 0.1040 \\
\midrule
\textbf{Ours}
  & \cellcolor{orange!25}26.04 & \cellcolor{orange!25}0.9490 & \cellcolor{red!25}0.0401
  & \cellcolor{orange!25}28.36 & \cellcolor{orange!25}0.9707 & \cellcolor{red!25}0.0247
  & \cellcolor{orange!25}25.37 & \cellcolor{orange!25}0.9506 & \cellcolor{red!25}0.0457
  & \cellcolor{orange!25}24.55 & \cellcolor{orange!25}0.9234 & \cellcolor{red!25}0.0651 \\
\textbf{Ours}(MSE) 
  & \cellcolor{red!25}26.60 & \cellcolor{red!25}0.9525 & \cellcolor{orange!25}0.0482
  & \cellcolor{red!25}28.82 & \cellcolor{red!25}0.9721 & \cellcolor{orange!25}0.0256
  & \cellcolor{red!25}25.66 & \cellcolor{red!25}0.9538 & \cellcolor{orange!25}0.0560
  & \cellcolor{red!25}25.11 & \cellcolor{red!25}0.9286 & \cellcolor{orange!25}0.0774\\
\bottomrule
\end{tabular}%
}
\end{table}

\begin{table}[t]
\centering
\caption{\textbf{Self-enact metrics per person.} We report novel-pose quantitative objective metrics separately for each person used in the evaluation on the Decaf dataset.}
\label{tab:novel_pose_supp_matt}
\resizebox{\textwidth}{!}{%
\begin{tabular}{l ccc ccc ccc ccc}
\toprule
& \multicolumn{3}{c}{S1} & \multicolumn{3}{c}{S2} & \multicolumn{3}{c}{S3} & \multicolumn{3}{c}{S4} \\
\cmidrule(lr){2-4} \cmidrule(lr){5-7} \cmidrule(lr){8-10} \cmidrule(lr){11-13}
Method & PSNR$\uparrow$ & SSIM$\uparrow$ & LPIPS$\downarrow$ & PSNR$\uparrow$ & SSIM$\uparrow$ & LPIPS$\downarrow$ & PSNR$\uparrow$ & SSIM$\uparrow$ & LPIPS$\downarrow$ & PSNR$\uparrow$ & SSIM$\uparrow$ & LPIPS$\downarrow$ \\
\midrule
Splatt. Av.~\cite{shao2024splattingavatar}
  & 25.59 & 0.9443 & 0.0834
  & 23.61 & 0.9313 & 0.0939
  & 25.68 & 0.9490 & 0.0730
  & 24.42 & 0.9353 & 0.0886 \\
Gauss. Av.~\cite{qian2024gaussianavatars}
  & 26.36 & 0.9556 & 0.0614
  & \cellcolor{red!25}23.68 & \cellcolor{red!25}0.9357 & \cellcolor{orange!25}0.0766
  & 25.85 & 0.9556 & 0.0570
  & 24.67 & 0.9447 & 0.0703 \\
\midrule
\textbf{Ours}
  & \cellcolor{orange!25}27.12 & \cellcolor{orange!25}0.9611 & \cellcolor{red!25}0.0324
  & 21.99 & 0.9287 & 0.0782
  & \cellcolor{orange!25}26.35 & \cellcolor{orange!25}0.9591 & \cellcolor{red!25}0.0370
  & \cellcolor{orange!25}25.56 & \cellcolor{orange!25}0.9498 & \cellcolor{red!25}0.0443 \\
\textbf{Ours}(MSE) 
  & \cellcolor{red!25}27.87 & \cellcolor{red!25}0.9646 & \cellcolor{orange!25}0.0416
  & \cellcolor{orange!25}22.10 & \cellcolor{orange!25}0.9311 & \cellcolor{red!25}0.0751
  & \cellcolor{red!25}26.81 & \cellcolor{red!25}0.9608 & \cellcolor{orange!25}0.0454
  & \cellcolor{red!25}26.13 & \cellcolor{red!25}0.9527 & \cellcolor{orange!25}0.0552\\
\bottomrule
\end{tabular}%
}
\end{table}

\clearpage
\section{Number of Channels}
We analyze the influence of the number of neural texture channels on rendering quality and inference speed. As shown in \Cref{tab:nt_channels}, increasing the number of channels from 3 to 6 significantly improves reconstruction quality, leading to higher PSNR and SSIM and lower LPIPS. Using only 3 channels limits the representation capacity of the neural textures, since the feature space must simultaneously encode RGB appearance as well as view- and pose-dependent effects such as color changes and local deformations caused by hand–face interaction. In our experiments, we increase the number of channels by doubling them at each step (3, 6, 12, 24). Beyond 6 channels, reconstruction quality saturates: 12 channels perform slightly worse than 6, while 24 channels bring only a marginal further gain, suggesting the additional capacity is not consistently exploited once the texture already captures the dominant appearance variations. Inference speed degrades only mildly up to 12 channels, but collapses sharply at 24 channels, as the per-splat memory footprint and channel-wise computation costs come to dominate the rasterization process. We therefore adopt 6 channels for all our experiments, as it offers the best trade-off between representation capacity and rendering efficiency.

\begin{table}[tb]
    \centering
    \caption{\textbf{Number of channels.} We analyzed the metrics and inference speed for Subject 2 from the Decaf dataset, and selected the optimal number of channels for Neural Textures.}
    \label{tab:nt_channels}
    \begin{tabular}{l c c c c}
        \toprule
        Channels & PSNR$\uparrow$ & SSIM$\uparrow$ & LPIPS$\downarrow$ & FPS$\uparrow$ \\
        \midrule
        3 &  27.54 & 0.9671 & 0.0286 & 153 \\
        \rowcolor{gray!20}
        6 & 28.12 & 0.9691 & 0.0259 & 149 \\
        12 & 27.89 & 0.9686 & 0.0276 & 144 \\
        24 & 28.22 & 0.9701 & 0.0270 & 27 \\
        %\bottomrule
    \end{tabular}
\end{table}
%\end{comment}

\end{document}